\documentclass[10pt,journal,compsoc]{IEEEtran}

\usepackage{amsmath,amsfonts}
\usepackage{algorithmic}
\usepackage{array}
\usepackage[caption=false,font=normalsize,labelfont=sf,textfont=sf]{subfig}
\usepackage{textcomp}
\usepackage{stfloats}
\usepackage{url}
\usepackage{verbatim}
\usepackage{graphicx}
\usepackage{graphicx} 
\usepackage{amsmath}
\usepackage{amsfonts}
\usepackage{multirow}
\usepackage{color,xcolor}
\usepackage{bbding}
\usepackage{graphicx}
\usepackage{amsmath}
\usepackage{amssymb}
\usepackage{booktabs}

\usepackage{enumitem}
\usepackage{multirow}

\usepackage{algorithm}
\usepackage{algorithmic}

\usepackage{newfloat}
\usepackage{listings}
\DeclareCaptionStyle{ruled}{labelfont=normalfont,labelsep=colon,strut=off} 
\floatstyle{ruled}
\newfloat{listing}{tb}{lst}{}
\floatname{listing}{Listing}

\usepackage[colorlinks,
            linkcolor=red,       
            anchorcolor=blue,  
            citecolor=blue,        
            ]{hyperref}

\usepackage{amsthm} 

\newtheorem{definition}{Definition}

\usepackage{orcidlink}

\ifCLASSOPTIONcompsoc
  \usepackage[nocompress]{cite}
\else
  \usepackage{cite}
\fi
\ifCLASSINFOpdf
\else
\fi

\usepackage{enumitem}
\begin{document}
%
\title{Data-Centric Long-Tailed Image Recognition}
%
%
%
%
\author{Yanbiao Ma~\orcidlink{0000-0002-8472-1475},
        Licheng Jiao~\orcidlink{0000-0003-3354-9617},~\IEEEmembership{Fellow,~IEEE,}
        Fang Liu~\orcidlink{0000-0002-5669-9354},~\IEEEmembership{Senior Member,~IEEE,}
        Shuyuan Yang~\orcidlink{0000-0002-4796-5737},~\IEEEmembership{Senior Member,~IEEE,}
        Xu Liu~\orcidlink{0000-0002-8780-5455},~\IEEEmembership{Member,~IEEE,}
        Puhua Chen~\orcidlink{0000-0001-5472-1426},~\IEEEmembership{Senior Member,~IEEE}
        }
\IEEEtitleabstractindextext{%
\begin{abstract}
In the context of the long-tail scenario, models exhibit a strong demand for high-quality data. Data-centric approaches aim to enhance both the quantity and quality of data to improve model performance. Among these approaches, information augmentation has been progressively introduced as a crucial category. It achieves a balance in model performance by augmenting the richness and quantity of samples in the tail classes. However, there is currently a lack of research into the underlying mechanisms explaining the effectiveness of information augmentation methods. Consequently, the utilization of information augmentation in long-tail recognition tasks relies heavily on empirical and intricate fine-tuning.
This work makes two primary contributions. Firstly, we approach the problem from the perspectives of feature diversity and distribution shift, introducing the concept of Feature Diversity Gain (FDG) to elucidate why information augmentation is effective. We find that the performance of information augmentation can be explained by FDG, and its performance peaks when FDG achieves an appropriate balance. Experimental results demonstrate that by using FDG to select augmented data, we can further enhance model performance without the need for any modifications to the model's architecture. Thus, data-centric approaches hold significant potential in the field of long-tail recognition, beyond the development of new model structures.
Furthermore, we systematically introduce the core components and fundamental tasks of a data-centric long-tail learning framework for the first time. These core components guide the implementation and deployment of the system, while the corresponding fundamental tasks refine and expand the research area.
\end{abstract}

\begin{IEEEkeywords}
Representational learning, Long-Tailed Recognition, Image classification, Data-Centirc AI.
\end{IEEEkeywords}}

\maketitle

\IEEEdisplaynontitleabstractindextext

%
\IEEEpeerreviewmaketitle

\ifCLASSOPTIONcompsoc
\IEEEraisesectionheading{\section{Introduction}\label{sec:introduction}}
\else
\section{Introduction}
\label{sec:introduction}
\fi

%
%
%
%

\begin{figure}[!t]
\centering
\includegraphics[width=3.5in]{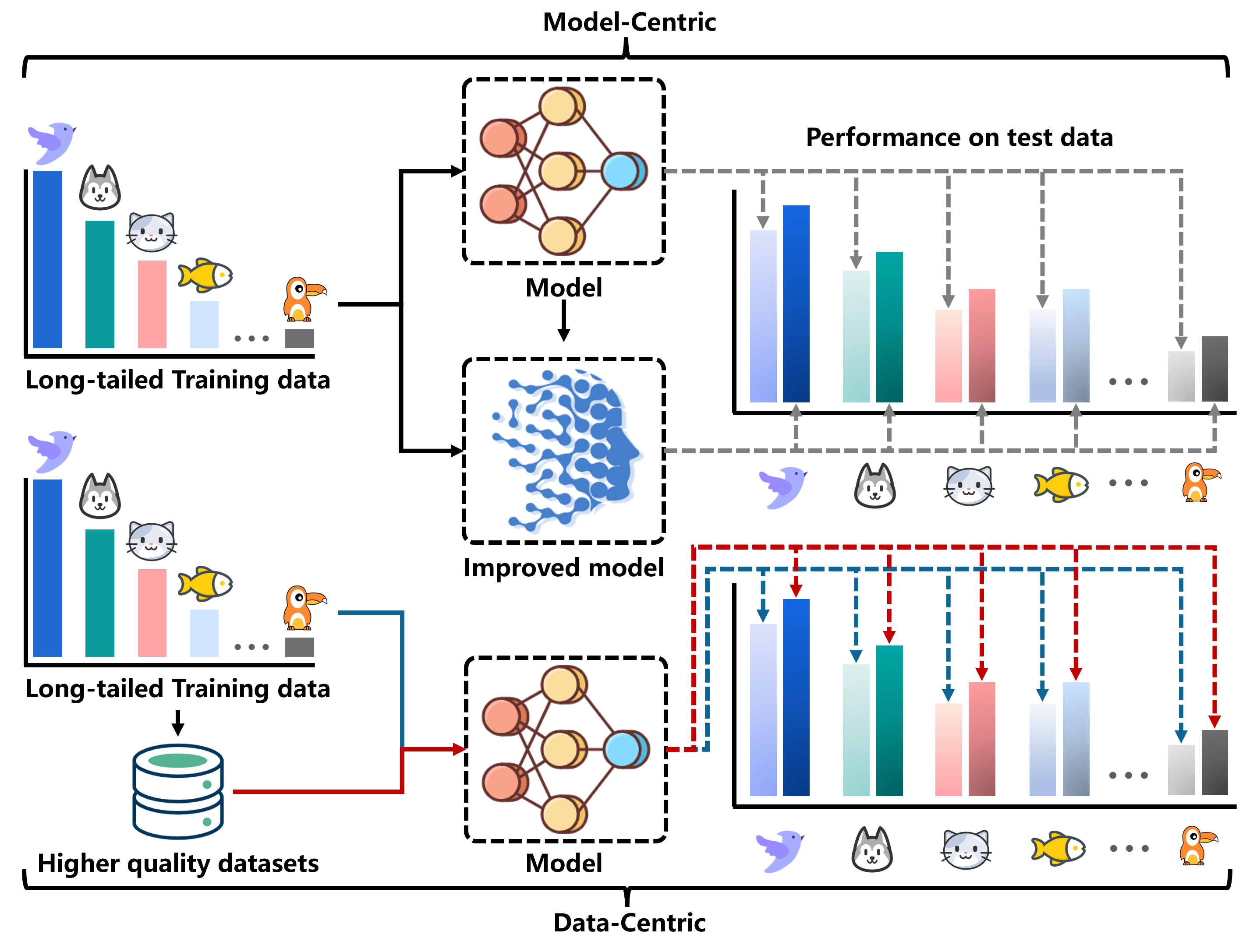}
\vskip -0.1in
\caption{In the traditional development cycle of AI models, researchers primarily focus on improving the model's architecture or training techniques to enhance the performance of long-tailed recognition. The common research paradigm involves assessing performance differences between different models, given training and test data. Data-centric long-tailed learning, on the other hand, should keep the model constant and concentrate on improving the quality of the dataset.}
\label{fig0}
\vskip -0.05in
\end{figure}

\IEEEPARstart{L}{ong-tailed} distribution of data is widely encountered in practical applications, and the imbalance in data distribution often leads to poor model performance in certain categories \cite{paper94,paper55}. To address this bias, a series of long-tailed learning methods have emerged, primarily categorized into two types: model-centric approaches and data-centric approaches. As shown in Fig.\ref{fig0}, model-centric approaches aim to enhance the model's structure to improve long-tailed recognition performance. Examples include Decoupled Training \cite{paper41,paper17,paper2,paper43} and Reweighted Loss \cite{paper29,paper31,paper8,paper23,paper34,paper36,paper38,paper58}, among others. In contrast, data-centric information augmentation focuses on improving the quality of training data by augmenting samples from the tail categories, thereby enhancing model performance. This includes data augmentation techniques such as Mixup, Cutout, and CutMix, as well as knowledge-guided tail class augmentation methods like OFA\cite{paper5}, CMO\cite{paper33}, and FDC\cite{paper63}. While past research has predominantly focused on improving the structure of long-tail recognition models, the benefits of model structure enhancements have gradually diminished with the rapid development of base models \cite{paper66,paper68}. Consequently, data-centric long-tailed recognition has become increasingly important \cite{paper70,paper71,paper72}. However, these methods still face two challenges:
\begin{itemize}[]
\item[(1)] The effectiveness of information augmentation methods often relies on empirical knowledge rather than quantitative metrics \cite{paper65}.
\item[(2)] Data-centric long-tailed recognition lacks a systematic framework \cite{paper69,paper73}. Optimizing data quality is a system engineering task that requires collaboration across multiple fundamental tasks, beyond just relying on information augmentation.
\end{itemize}

Different information augmentation methods exhibit significant variations in their performance when dealing with different scenarios. Traditional data augmentation methods tend to perform well when tail-class samples sufficiently cover their true distribution (Definition \ref{def2}). However, in practice, tail-class samples are often limited and cannot adequately cover their true distribution, making traditional data augmentation methods less effective \cite{paper5,paper63}. Hence, we need to introduce additional knowledge to augment tail-class samples to facilitate the model's learning of potential distributions (Definition \ref{def3}). For example, the FDC utilizes the variance of the head-class distribution to re-estimate the tail-class distribution, while CMO and OFA combine the background information from head-class samples with the foreground information from tail-class samples to generate richer tail-class samples. However, the current use of information augmentation methods often heavily relies on empirical fine-tuning of parameters, lacking quantified research on how to generate and select appropriate augmented samples. When augmented samples deviate beyond the true distribution range, they may conflict with sample labels, disrupting the model's learning process \cite{paper11,paper29}. Conversely, when augmented samples still fall within the observed distribution (Definition \ref{def1}), they may fail to provide new information for tail classes. To address this challenge, we propose a model-agnostic metric, known as Feature Diversity Gain (as detailed in Section \ref{sec3}), for assessing and selecting suitable augmented samples. This metric offers a robust tool and guidance for the use of information augmentation methods. Experimental results indicate that when the FDG (Feature Diversity Gain) values of augmented samples fall within a reasonable range, information augmentation methods can achieve their optimal performance (Section \ref{sec4}).

The core concept of data-centric approaches aims to enhance the performance of AI models by improving the quality of the dataset. However, when dealing with long-tailed data, relying solely on information augmentation strategies is not sufficient. Directly performing data augmentation before data cleaning may amplify errors \cite{paper108}, which can be disastrous for long-tail recognition models \cite{paper64,paper62}. Therefore, to address this challenge, our goal is to propose a systematic data processing framework specifically tailored to long-tailed recognition tasks (Section \ref{sec5}). This framework includes multiple fundamental tasks and components, such as assessing various types of data imbalance and identifying mislabeled data. Through this work, we aspire to advance the role of data-centric long-tailed learning methods in tackling real-world imbalanced data challenges.

\section{RELATED WORK}\label{sec2}

\subsection{Class Rebalancing}\label{sec2.1}

Real-world data typically exhibit long-tailed distributions, which leads to models that do not accurately learn the true distribution of tail classes. The purpose of resampling \cite{paper45,paper54,paper10,paper47} is to rebalance the distribution of the data during training so that the losses incurred by each class are relatively balanced. The idea of reweighting is more intuitive, and it alleviates the model’s over-focus on the head classes by directly balancing the losses \cite{paper8,paper23,paper29,paper34,paper2,paper36,paper38}. Previous reweighting and resampling strategies have focused on enhancing tail classes, but some recent studies have observed that tail classes are not always difficult to learn. \cite{paper37} utilizes a nonlinear transformation of the class accuracy to compute the weights of classes. DSB \cite{paper29} and DCR \cite{paper31}, for the first time, examined the factors influencing model bias from a geometric perspective and proposed a rebalancing approach.

Rebalancing strategies are also commonly used in conjunction with decoupled training to rebalance the classifier. For example, \cite{paper17} utilizes resampling strategies to fine-tune the classifier in the second phase of decoupled training. \cite{paper2} proposes to learn the classifier with class-balanced loss by adjusting the weight decay and MaxNorm in the second stage. BBN \cite{paper61} designs a two-branch network to rebalance the classifier, which is consistent with the idea of decoupled training. To avoid the damage of decoupled training on the performance of head classes, SimCal \cite{paper43} trains networks with dual branches, one of which is used to rebalance the classifier while the other is used to maintain the performance of head classes. Some other approaches (GCL \cite{paper21}, ECM \cite{paper16}, ALA \cite{paper59}) calibrate the data distribution by adjusting the logit to have larger margins between classes.

\subsection{Information Augmentation}\label{sec2.2}

When the few samples of the tail class do not represent its true distribution well, the model cannot learn information outside the observed distribution \cite{paper5,paper63}, no matter how one goes about adjusting the sampling strategy and balancing the losses. Information augmentation facilitates imbalance learning by introducing additional knowledge, and can be divided into two types of methods: knowledge transfer and data augmentation.

Head-to-tail knowledge transfer aims to transfer knowledge from head class to tail class to improve the performance of tail class. FTL \cite{paper52}, FDC \cite{paper63} and LEAP \cite{paper26} generate enhancement samples by transferring head class variance to tail class. It's worth noting that FDC provides the experimental foundation for transfer variance for the first time. OFA \cite{paper5} decomposes the features of each class into class-general features and class-specific features. During training, the class-specific features of the tail class are fused with the generic features of the head class to generate new features to augment the tail class. GIST \cite{paper24} proposes to transfer the geometric information of the feature distribution boundaries of the head class to the tail class. The motivation of CMO \cite{paper33} is very intuitive; it considers that the images of the head class have a rich background, so the images of the tail class can be directly pasted onto the rich background of the head class to increase the richness of the tail class. CUDA \cite{paper1} gives the appropriate enhancement intensity for each class through empirical studies, which can be combined with other augmentation methods.

Data augmentation in long-tailed recognition improves the performance of tail classes by improving conventional data augmentation methods. MiSLAS \cite{paper60} and Remix \cite{paper4} suggest employing mixup to enhance feature learning. \cite{paper56} discovers that mixup has good accuracy when combined with resampling. FASA \cite{paper54} proposes generating features based on Gaussian priors and evaluating weak classes on balanced datasets to adjust the sampling rate. MetaSAug \cite{paper22} generates enhanced features for tail classes with ISDA. Recent studies show that noise-based data augmentation can also play an important role in long-tail recognition. Similar to adversarial attacks, M2m \cite{paper18}, AdvProp \cite{paper48} and FedAFA \cite{paper28} propose to transform some of the head class samples into tail class samples by perturbation-based optimization to achieve tail class augmentation. Unlike adversarial noise, OpeN \cite{paper53} attempts to enhance tail classes directly using pure noise images, which outperforms relative to the adversarial noise-based approach.

Unlike the methods mentioned above, OUR \cite{paper112} identified and defined a long-tailed phenomenon of model robustness, expanding the research scope of long-tailed learning. Furthermore, OUR introduced orthogonal uncertainty representations of manifolds for augmenting tail classes, significantly enhancing the model's robustness on tail classes.

\section{Data-Centric Performance Metrics}
\label{sec3}

We aim to construct a measure solely dependent on data to explain the effectiveness of information augmentation and enhance the quality of data using data-centric measures to further unleash the performance of existing methods. The content of this section is organized as follows: We first discuss, from the perspective of data manifold, the potential mechanisms related to the performance of information augmentation methods. Then, we propose a measure for quantifying the increase in diversity brought about by information augmentation methods. It is important to note that our measure is calculated directly from the data distribution and does not depend on the model.

\subsection{Inspiration from the Marginal Effects of Manifold}
\label{sec3.1}

We begin by defining the concepts of observed distribution, underlying distribution, and true distribution.

\begin{definition}[Observed Distribution]
\label{def1}
The distribution comprised of available samples (i.e., the training set).
\end{definition}

\begin{definition}[True Distribution]
\label{def2}
The distribution formed by samples when there are enough samples of a class to fully encompass all the characteristics of that class.
\end{definition}

\begin{definition}[Underlying distribution]
\label{def3}
The true distribution other than the observed distribution.
\end{definition}

The Law of Manifold Distribution \cite{paper19,paper31} postulates that a specific category of natural data resides in proximity to a low-dimensional manifold embedded within a high-dimensional space, while samples from different categories are distributed around distinct manifolds. \cite{paper29}, by utilizing the volume of the manifold, quantified the diversity of category features and pointed out that decision boundaries tend to favor categories with smaller volumes, leading to model bias. However, in long-tail scenarios, the observed distribution of tail categories often fails to sufficiently cover their true distribution. To address this issue, information augmentation methods enhance the diversity of tail class samples, expanding the volume of their observed distribution. This allows the model to learn more information from outside the training domain. Therefore, we reasonably infer that \textbf{the diversity gain brought by information augmentation to tail classes could be a potential mechanism underlying its effectiveness}.

The marginal effect \cite{paper8} suggests that the diversity of features for each category is limited, meaning that data manifolds have boundaries, and samples beyond these boundaries do not exist in the real world (As shown in Fig.\ref{fig1}A). Information augmentation methods might generate samples that lie outside the scope of the true distribution. When newly generated samples deviate significantly from the observed distribution, it can result in overlapping between the data manifold of tail classes and other manifolds, thus compromising the model's performance \cite{paper63}. Consequently, \textbf{we are interested in investigating the impact of the deviation between the augmented samples generated by information augmentation methods and the observed distribution on model performance}. In the following sections, we will strive to introduce a metric that can simultaneously capture the gain in diversity and the shift in distribution.

\subsection{Feature Diversity Gain}
\label{sec3.2}

Information augmentation methods can generate augmented samples both in the image space and in the embedded space. To maintain the generality of the proposed metric, we uniformly assume the dimensionality of the samples to be $d$. Below, we employ the volume of the data manifold to measure category diversity and subsequently define the Feature Diversity Gain.

\begin{figure}[h]
\centering
\includegraphics[width=3.5in]{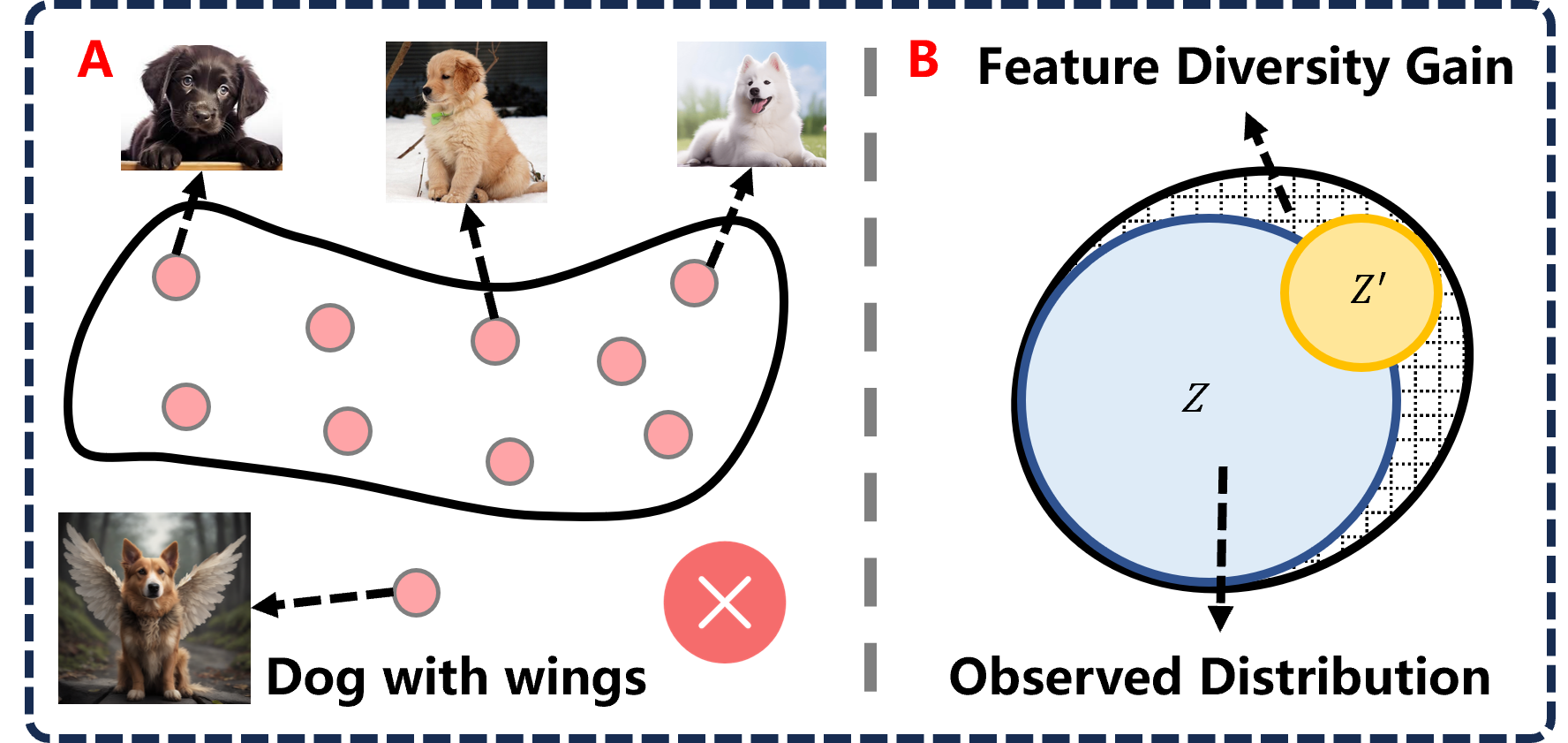}
\vskip -0.05in
\caption{A: The data manifold corresponding to a class has boundaries, and samples outside the manifold boundaries may not exist in the real world, such as dogs with long wings. B: Grid regions represent the diversity gain in features brought about by augmented samples.}
\label{fig1}
\end{figure}

Given a class of data $X=[x_1,x_2,\dots,x_k]\in \mathbb{R}^{d \times N}$, where $d$ is the dimensionality of the samples and $N$ is the number of samples in that category. Firstly, estimate the covariance matrix $\Sigma=\mathbb{E}[\frac{1}{N} {\textstyle \sum_{i=1}^{N}}x_ix_i^T ]=\frac{1}{N}XX^T\in \mathbb{R}^{d\times d}$ of the sample set $X$. Next, the volume of the subspace spanned by the sample set $X$ can be succinctly expressed as $\sqrt{\det(\frac{1}{N}XX^T)}$, which can be obtained through the singular value decomposition of matrix $X$. To address situations where $\frac{1}{N}XX^T$ is not full rank and leads to a volume of $0$, an identity matrix is added to the covariance matrix, without affecting the monotonicity of the volume. Furthermore, for enhanced numerical stability, the volume of the data manifold corresponding to the logarithmically transformed sample set $X$ can be calculated as
\begin{equation}
\begin{split}
V(X)=\frac{1}{2}\log_2\det(I+\frac{1}{N}XX^T).
\nonumber
\end{split}
\end{equation}

The computed result from the above equation can be regarded as a measure of feature diversity for the sample set $X$. Based on this, we introduce the definition of Feature Diversity Gain (FDG). The Feature Diversity Gain represents the increment in data diversity brought about by using information augmentation methods for a specific category. Moreover, we aim for this metric to also reflect the extent of distribution shift between augmented samples and the original samples. In the image space, the diversity of the sample set remains fixed. However, in the embedded space, we observe that different models extracting embeddings for the same sample set exhibit uncertain magnitudes of feature diversity. This uncertainty may arise due to the stochasticity of parameter initialization. Considering that this study involves multiple sets of experiments to observe the relationship between Feature Diversity Gain and the performance of information augmentation methods, we desire the quantified results of Feature Diversity Gain to remain unaffected by the magnitude of feature diversity.

Building upon the aforementioned motivation, we define the Feature Diversity Gain in a ratio-based manner. Specifically, the Feature Diversity Gain can be measured as the difference between the feature diversity of the augmented dataset and the feature diversity of the initial samples, relative to the feature diversity of the initial samples. For a more intuitive explanation, we utilize the blue area in Fig.\ref{fig1}B to represent the distribution of the initial sample $Z$, the yellow area for the distribution of augmented samples, and the gridded area to indicate the increased feature diversity resulting from applying information augmentation techniques. The proposed metric should adhere to the following principles:
\begin{itemize}
\item \textbf{Principle 1}: When the count of augmented samples is $0$, the Feature Diversity Gain is $0$.

\item \textbf{Principle 2:} Regardless of the count of augmented samples, when all augmented samples cluster at a single point, they do not provide additional information and might even increase training complexity. Therefore, we expect the Feature Diversity Gain to be negative in this scenario and reach its minimum value (lower bound).
\end{itemize}

Based on the above principles, we formally define the measure of feature diversity gain as follows.

\begin{definition}[\textbf{Feature Diversity Gain}]
Let $A$ be an information augmentation method, and $X'$ be the augmented samples generated from $X$ using method $A$, with a count of $N'$. Let $M$ denote the model trained using both $X'$ and $X$, and $Z'$ and $Z$ represent the d-dimensional features obtained from $M$ corresponding to $X'$ and $X$, respectively. Let $F=[Z,Z']\in \mathbb{R}^{d\times (N+N')}$, and perform mean normalization on $Z$, $Z'$, and $F$. We define the Feature Diversity Gain (FDG) after augmentation using method $A$ as
\begin{equation}
\begin{split}
&FDG(A,F,Z)=\frac{V(F)-V(Z)}{V(Z)} \\
&=\frac{\log_2\det(I+\frac{1}{N+N'}FF^T)-\log_2\det(I+\frac{1}{N}ZZ^T)}{\log_2\det(I+\frac{1}{N}ZZ^T)} \\
&=\frac{\log_2\det(I+\frac{1}{N+N'}FF^T)^{-1}\det(I+\frac{1}{N}ZZ^T)}{\log_2\det(I+\frac{1}{N}ZZ^T)} \\
&=\log_\delta \frac{\det(I+\frac{1}{N+N'}FF^T)}{\det(I+\frac{1}{N}ZZ^T)}, \\
\delta &= \det(I+\frac{1}{N}ZZ^T),
\nonumber
\end{split}
\end{equation}
where $V(F)$ denotes the volume of $F$ and $V(Z)$ denotes the volume of $Z$. In the following, we verify that the FDG satisfies the two principles one by one.
\end{definition}

\begin{itemize}
\item $FDG(A,F,Z)$ can be expressed as $$\log_\delta \frac{\det(I+\frac{1}{N+N'}(ZZ^T+Z'Z'^T))}{\det(I+\frac{1}{N}ZZ^T)}.$$ When $N'=0$, $Z'Z'^T$ is a zero matrix, so $FDG(A,F,Z)\!=\!log_\delta 1\!=\!0$, which satisfies Principle 1.

\item We prove that when all augmented samples converge to a single point, $FDG$ reaches the lower bound $-\frac{N'}{N+N'}$, thus fulfilling Principle 2. The proof is outlined below.
\end{itemize}

\begin{proof}
The function $\log \det(\cdot)$ is strictly concave and satisfies the inequality: $\log\det( {\textstyle \sum_{j=1}^{k}}\alpha_jS_j)\ge  {\textstyle \sum_{j=1}^{k}}\alpha_j\log\det(S_j)$, where $\alpha_j>0$, $ {\textstyle \sum_{j=1}^{k}}\alpha_j=1,j=1,\dots,k$, and $S_j(j=1,\dots,k)$ are symmetric positive definite matrices. When $k=2$, we have $\log\det(\alpha_1S_1+\alpha_2S_2)\ge \alpha_1\log\det(S_1)+\alpha_2\log\det(S_2)$. By setting $\alpha_1=\frac{N}{N+N'}$, $\alpha_2=\frac{N'}{N+N'}$, $S_1=I+\frac{1}{N}ZZ^T$, and $S_2=I+\frac{1}{N'}Z'Z'^T$, we can derive
\begin{equation}
\begin{split}
\log\det(\alpha_1S_1\!+\!\alpha_2S_2)&=\log\det(I\!+\!\frac{1}{N+N'}(ZZ^T\!+\!Z'Z'^T)) \\
&=\log\det(I+\frac{1}{N+N'}FF^T).
\nonumber
\end{split}
\end{equation}
Thus, we have
\begin{equation}
\begin{split}
&\log\det(I+\frac{1}{N+N'}FF^T)\ge \frac{N}{N+N'}\log\det(I+\frac{1}{N}ZZ^T) \\
&+\frac{N'}{N+N'}\log\det(I+\frac{1}{N'}Z'Z'^T) \\
&\Rightarrow \frac{1}{2}\log\det(I+\frac{1}{N+N'}FF^T)\\
&\ge \frac{1}{N+N'}(\frac{N}{2}\log\det(I+\frac{1}{N}ZZ^T))\\
&+\frac{N'}{2}\log\det(I+\frac{1}{N'}Z'Z'^T) \\
&\Rightarrow V(F)\ge \frac{1}{N+N'}(NV(Z)+N'V(Z')).
\nonumber
\end{split}
\end{equation}
Further it can be obtained that
\begin{equation}
\begin{split}
\frac{V(F)-V(Z)}{V(Z)}\ge \frac{(\frac{N}{N+N'}-\frac{N+N'}{N+N'})V(Z)+\frac{N'}{N+N'}V(Z')}{V(Z)} \\
\Rightarrow \frac{V(F)-V(Z)}{V(Z)}\ge \frac{\frac{-N'}{N+N'}V(Z)+\frac{N'}{N+N'}V(Z')}{V(Z)}.
\nonumber
\end{split}
\end{equation}
Again, since $I+Z'$ is a positive definite matrix, $V(Z')\ge 0$. When the samples in $Z'$ are concentrated at one point, the $\frac{V(F)-V(Z)}{V(Z)}$ reaches a minimum (i.e., a lower bound) $\frac{-N'}{N+N'}$ since $V(Z')=0$.
\end{proof}

It's worth noting that \cite{paper65} assumes that the post-augmentation feature diversity is obtained by adding up the feature diversities of the initial samples and the augmented samples. However, we argue that the post-augmentation feature diversity should be measured by the volume of the subspace spanned jointly by the augmented and initial samples, rather than the sum of volumes spanned individually by each set. This is because we aim to avoid having the decision boundary separate the features of the same category. Additionally, another benefit of this approach is that it enables FDG to reflect the degree of offset between augmented and original samples.

\section{FDG Reflects the Performance of Information Augmentation Methods}
\label{sec4}

In this section, we will explore the relationship between FDG and the performance of information augmentation methods. As hypothesized and analyzed in Section \ref{sec3.1}, an information augmentation method is expected to achieve maximum performance when FDG falls within an appropriate range. When FDG is too low, it indicates that the additional information brought by the augmentation method is highly limited. On the other hand, when FDG is too high, it might lead to the generation of augmented samples that lack practical significance. Therefore, both excessively high and excessively low FDG values can constrain the performance of information augmentation methods.

\begin{figure*}[t]
\centering
\includegraphics[width=7.15in]{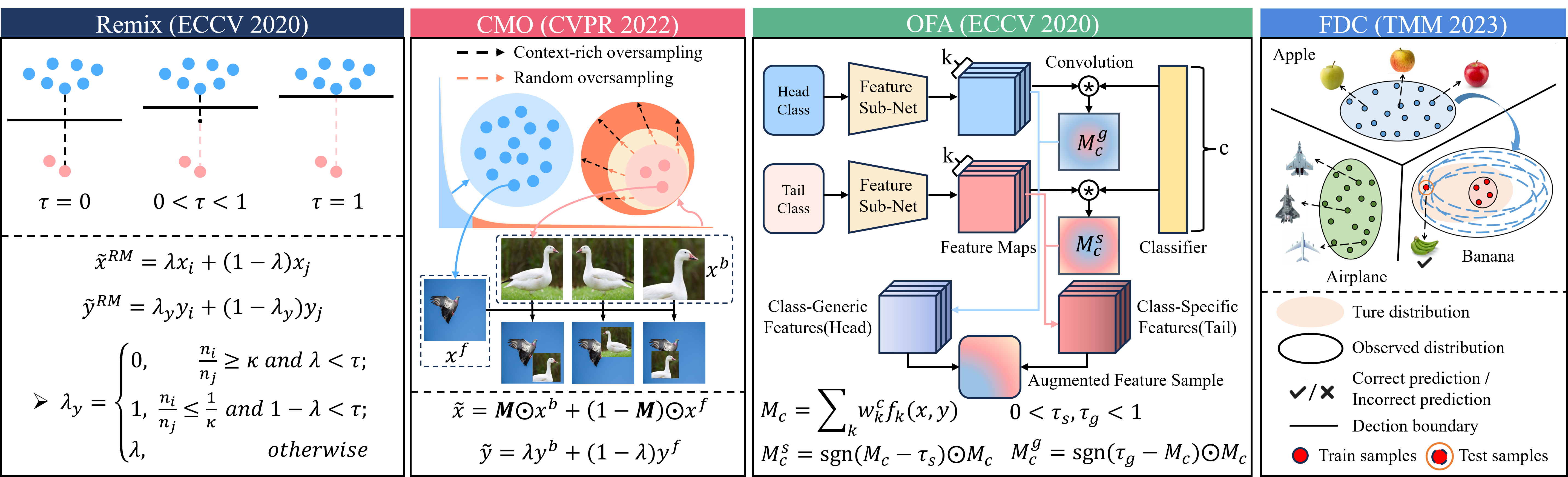}
\vskip -0.05in
\caption{\textbf{Remix} builds upon Mixup by refining the synthesized labels to ensure that when samples from the head class are mixed with samples from the tail class, the resulting label is entirely contributed by the tail class. \textbf{CMO} employs CutMix to patch segments from tail class samples onto head class samples, thereby generating augmented samples. \textbf{OFA} decomposes sample features into class-specific and class-generic features, and combines the tail class's specific features with the head class's generic features to create augmented samples for the tail class. \textbf{FDC} observes that similar classes have similar distribution statistics (variance), and thus, it re-estimates the distribution for the tail class by transferring the variance from the most similar head class, sampling augmented samples from the new distribution.}
\label{fig2}
\end{figure*}

To validate the above hypotheses, we conducted experiments on the long-tailed datasets CIFAR-10-LT and CIFAR-100-LT. Information augmentation can be performed in both the image space and the embedding space. In the image space, we selected the representative methods Remix \cite{paper4} and CMO \cite{paper33} for further investigation. In the embedding space, we opted for the advanced methods OFA \cite{paper5} and FDC \cite{paper63} as subjects of study. All four methods generate augmented samples for the tail classes to balance the training data. Fig.\ref{fig2} illustrates the core operations of these four information augmentation methods.

\subsection{How to Utilize Remix, CMO, OFA, and FDC}
\label{sec4.1}

Remix and CMO directly augment the tail classes in the image space and then proceed with training. OFA and FDC both employ decoupled training \cite{paper17}, thus the training phase consists of two parts. After the first phase concludes, OFA and FDC augment the tail classes in the embedding space and then feed the augmented samples into the second phase for training.

When using Remix, CMO, and OFA, it is necessary to select sample pairs to generate augmented samples. Assuming the number of samples in the tail class $t$ is $n_t$, if the sample count of a category exceeds $3n_t$, Remix considers that category as a relative head class compared to tail class $t$. Furthermore, Remix matches a head class sample for each sample from the tail class to create sample pairs. CMO directly samples background samples from the long-tailed distribution, while foreground samples are selected from the tail class. OFA matches a sample from the most similar head class for each tail class sample to create sample pairs. FDC controls the quantity of augmented samples by altering the sampling frequency.

Whether in the image space or the embedding space, our goal is to use the four methods mentioned above to balance the long-tailed dataset. Taking CIFAR-10-LT (IF=$100$) as an example, the class with the most samples has $5000$ samples, while the class with the fewest samples contains only $500$ samples. We need to generate $4500$ augmented samples for the class with the fewest samples in the tail. Similarly, the sample count for other tail classes also needs to be increased to $5000$. Here, we introduce the \cite{paper5} method for partitioning head and tail classes. Firstly, the categories are sorted in descending order of sample quantity, with the first $h$ categories selected as head classes and the rest as tail classes. The ratio of head class samples to the total number of samples is defined as $h_r\in (0,1)$, and $h$ is selected as the minimum value that satisfies $h_r>0.9$.

\subsection{Generating Augmented Data with Varying FDG}
\label{sec4.2}

To explore the impact of feature diversity gain brought by augmented data on model performance, it is necessary to generate multiple sets of augmented data with varying FDGs. Taking Remix as an example, a single Remix operation can generate sufficient augmented samples for the tail classes in a long-tail dataset. We define the overall feature diversity gain for tail classes as $FDG_{Tail}=\frac{1}{\mid Tail \ class\mid } { \sum_{t\in Tail \ class}} FDG_t$, where $FDG_t$ represents the feature diversity gain brought by the information augmentation method to tail class $t$, and $\mid Tail \ class\mid $ represents the number of tail classes. In this work, we used Remix $100$ times to generate $100$ different sets of augmented data and calculated the corresponding feature diversity gains. We sorted the $100$ FDGs in descending order and sampled $50$ of them uniformly. We saved the augmented data corresponding to these $50$ FDGs for experiments. However, there were relatively few instances where FDG values were either very small or very large among these $50$ FDGs.

To obtain multiple sets of augmented data with a wide range of feature diversity gains, we propose the following data selection process. Firstly, generate a sufficient number of augmented samples, and then cluster these augmented samples into multiple subsets using $k$-means clustering (set to $500$ subsets in this paper). Assuming each subset contains $10$ samples, if a tail class requires $450$ augmented samples, it would involve selecting $45$ subsets step by step. If augmented data with larger FDG values are needed, we require that at each step of subset selection, the chosen subset should maximize FDG compared to other subsets. Conversely, for augmented data with smaller FDG values, we aim to select subsets that minimize FDG at each step. By combining different subsets, augmented data with different FDGs can be generated. Following the above process, we generated $50$ sets of augmented data with either higher or lower FDGs, resulting in a total of $100$ sets of data. Subsequently, we retained $50$ sets of augmented data with uniformly distributed FDGs for experiments. Similarly, for each long-tailed dataset, CMO, OFA, and FDC each retained $50$ sets of augmented data.

\begin{figure*}[!t]
	\centering
	\begin{minipage}{0.243\linewidth}
		\centering
		\includegraphics[width=1\linewidth]{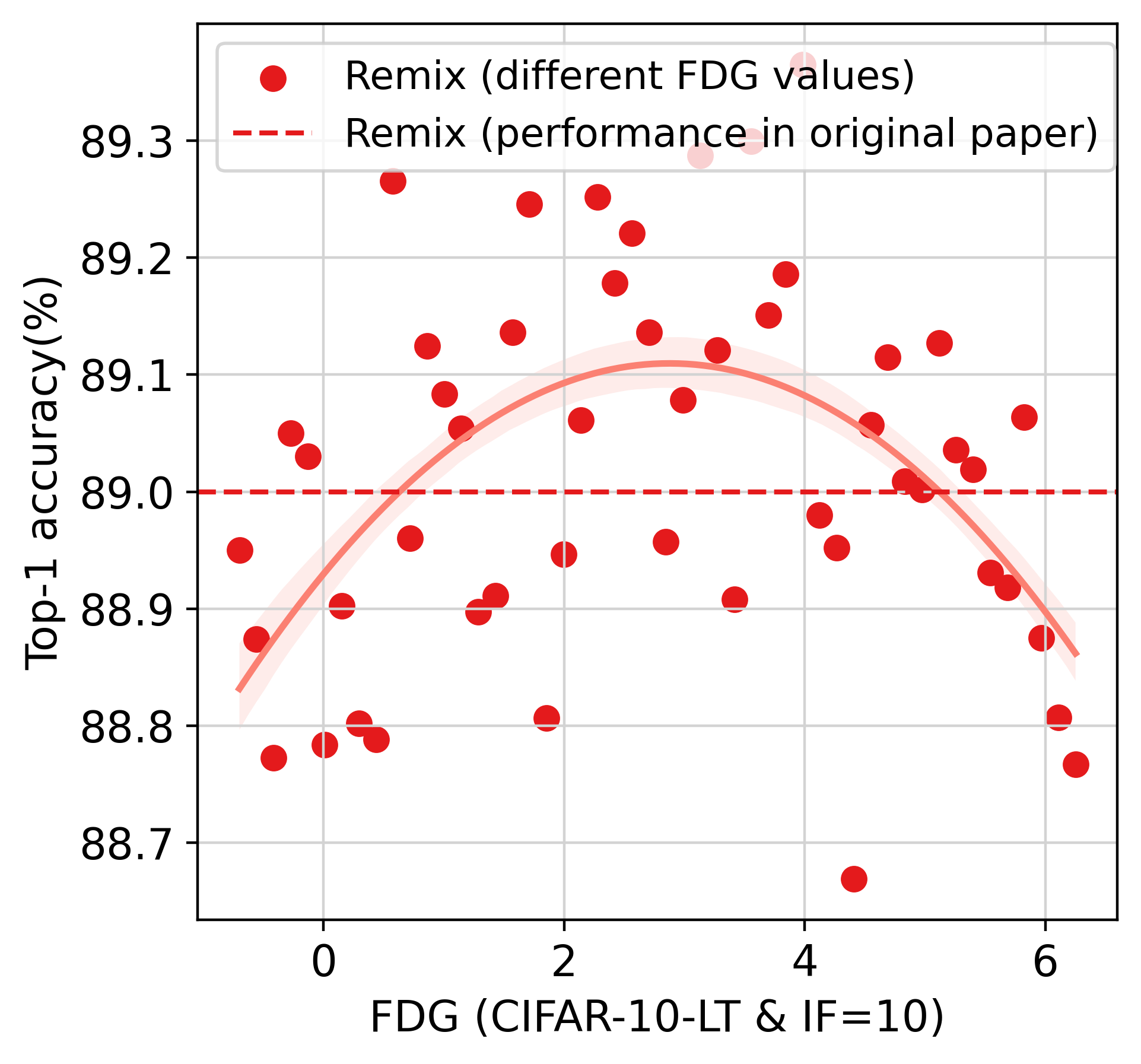}
	\end{minipage}
	\begin{minipage}{0.243\linewidth}
		\centering
		\includegraphics[width=1\linewidth]{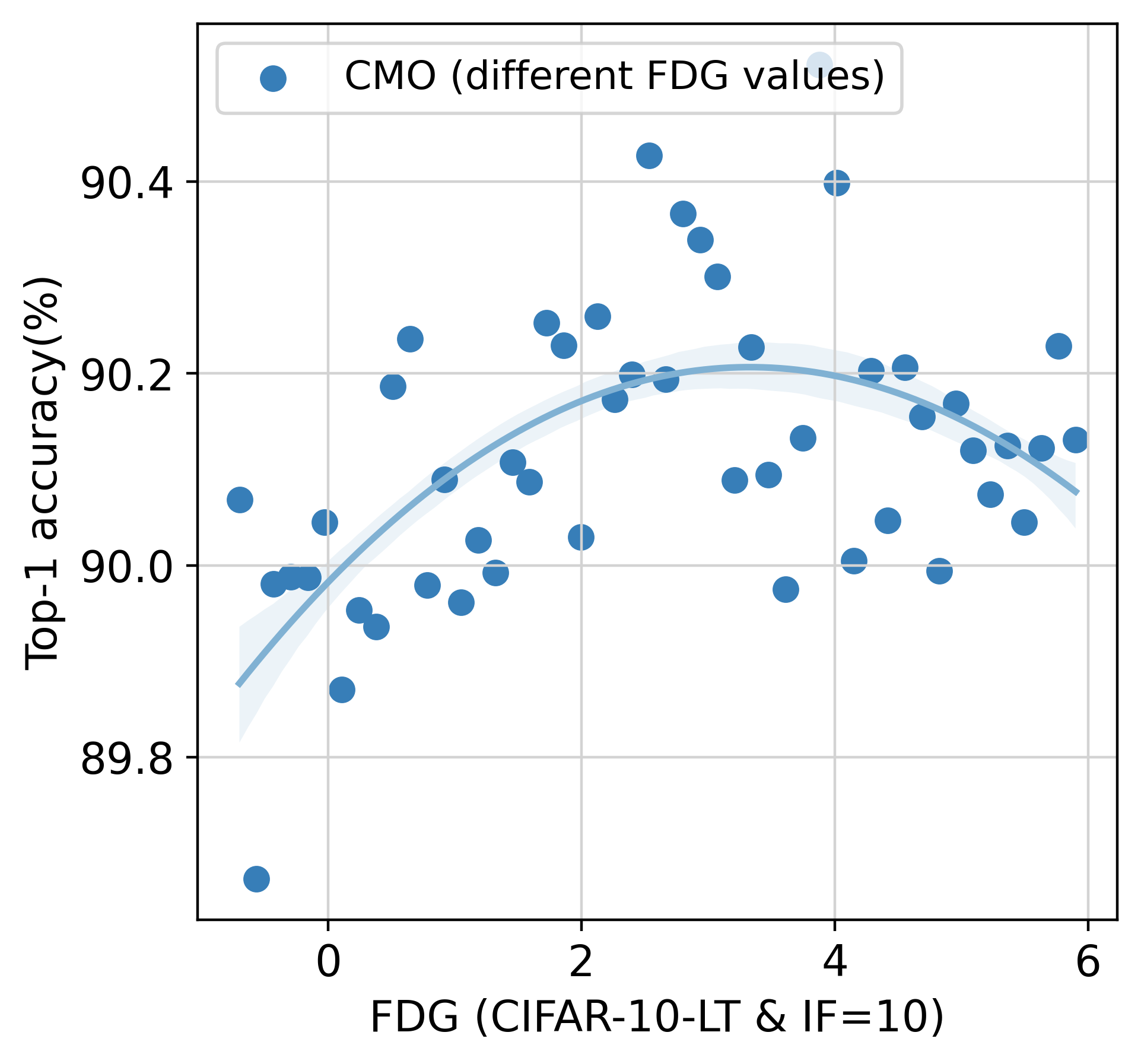}
	\end{minipage}
	\begin{minipage}{0.243\linewidth}
		\centering
		\includegraphics[width=1\linewidth]{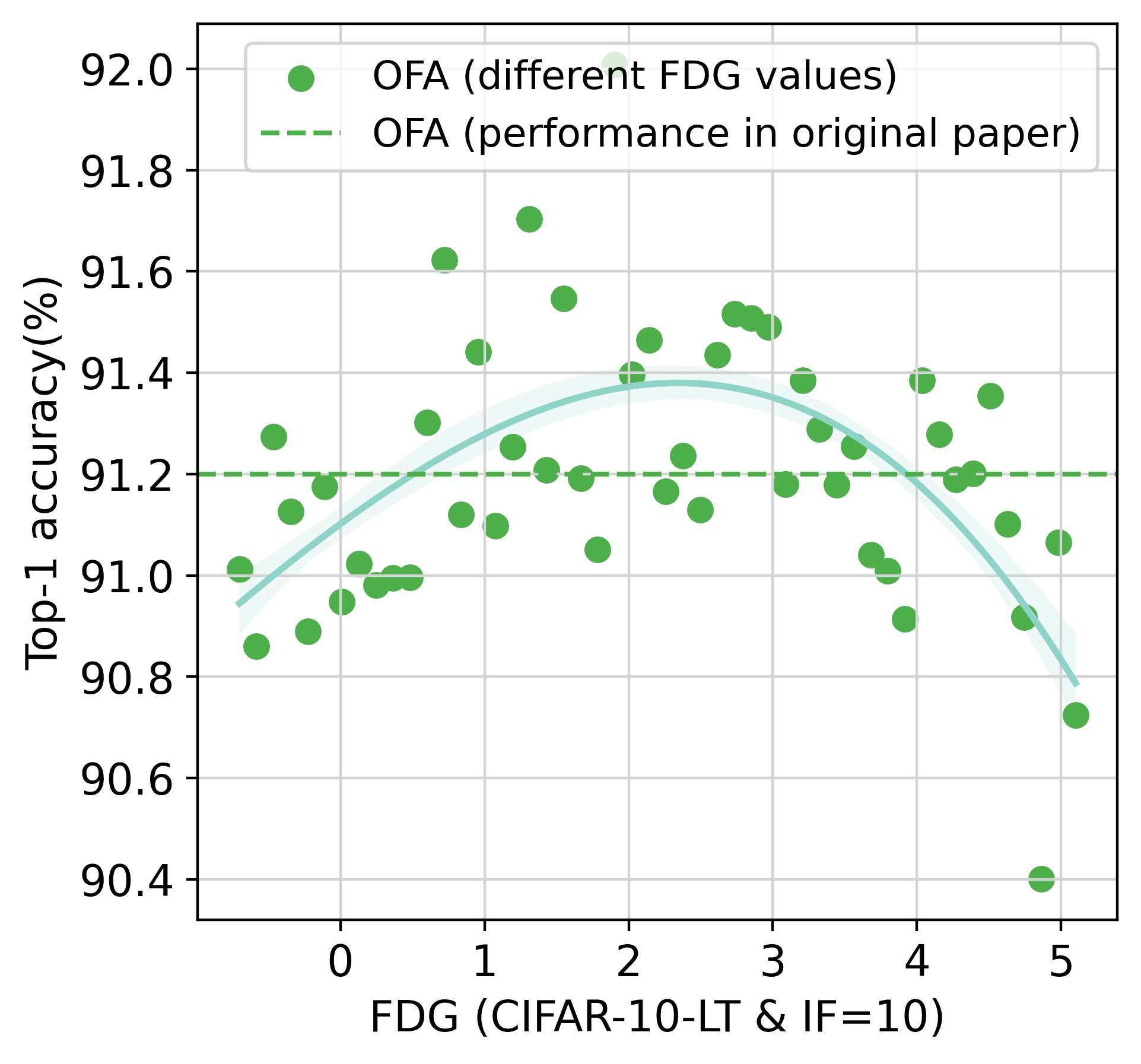}
	\end{minipage}
	\begin{minipage}{0.243\linewidth}
		\centering
		\includegraphics[width=1\linewidth]{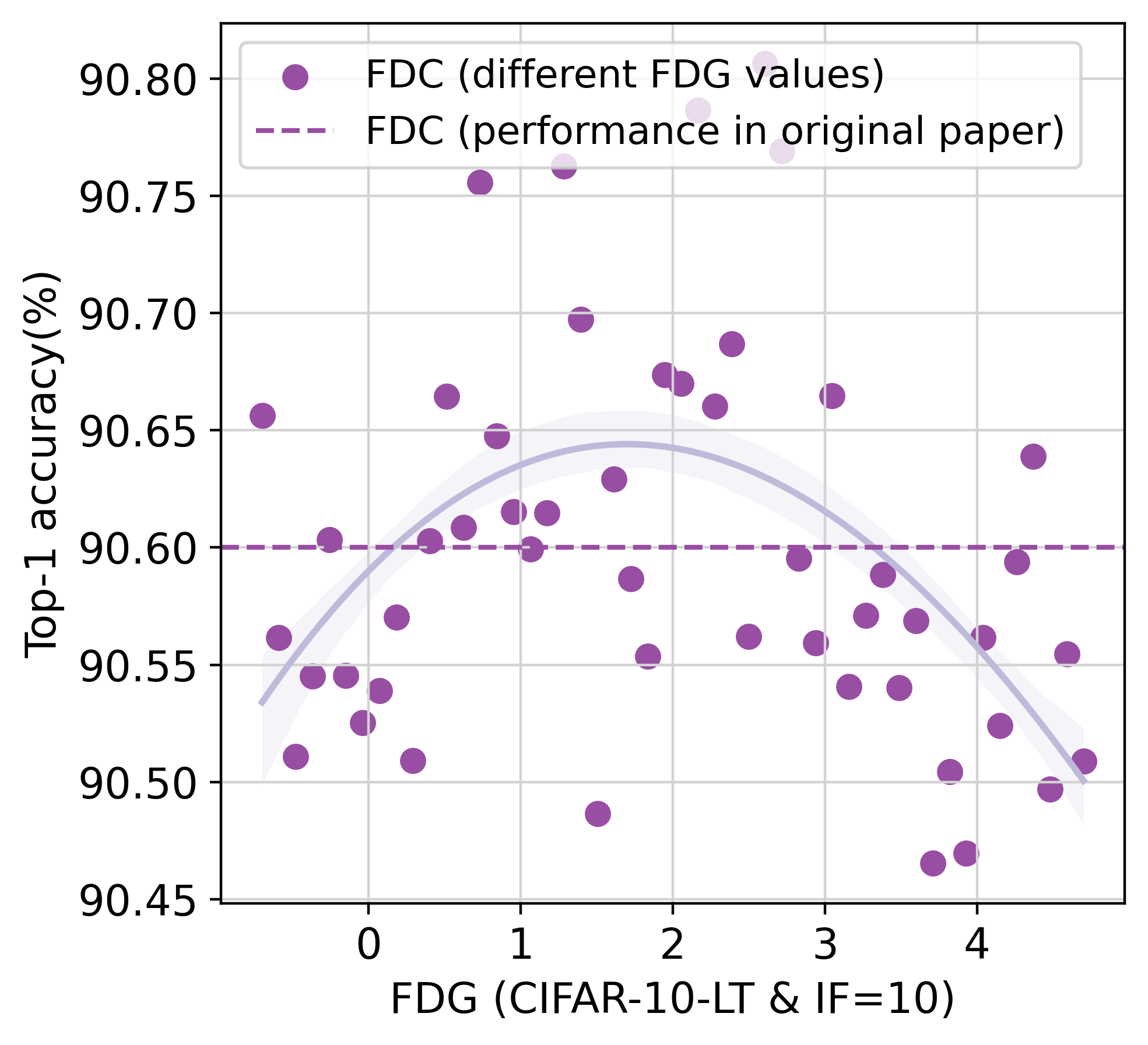}
	\end{minipage} 

     \centering
	\begin{minipage}{0.243\linewidth}
		\centering
		\includegraphics[width=1\linewidth]{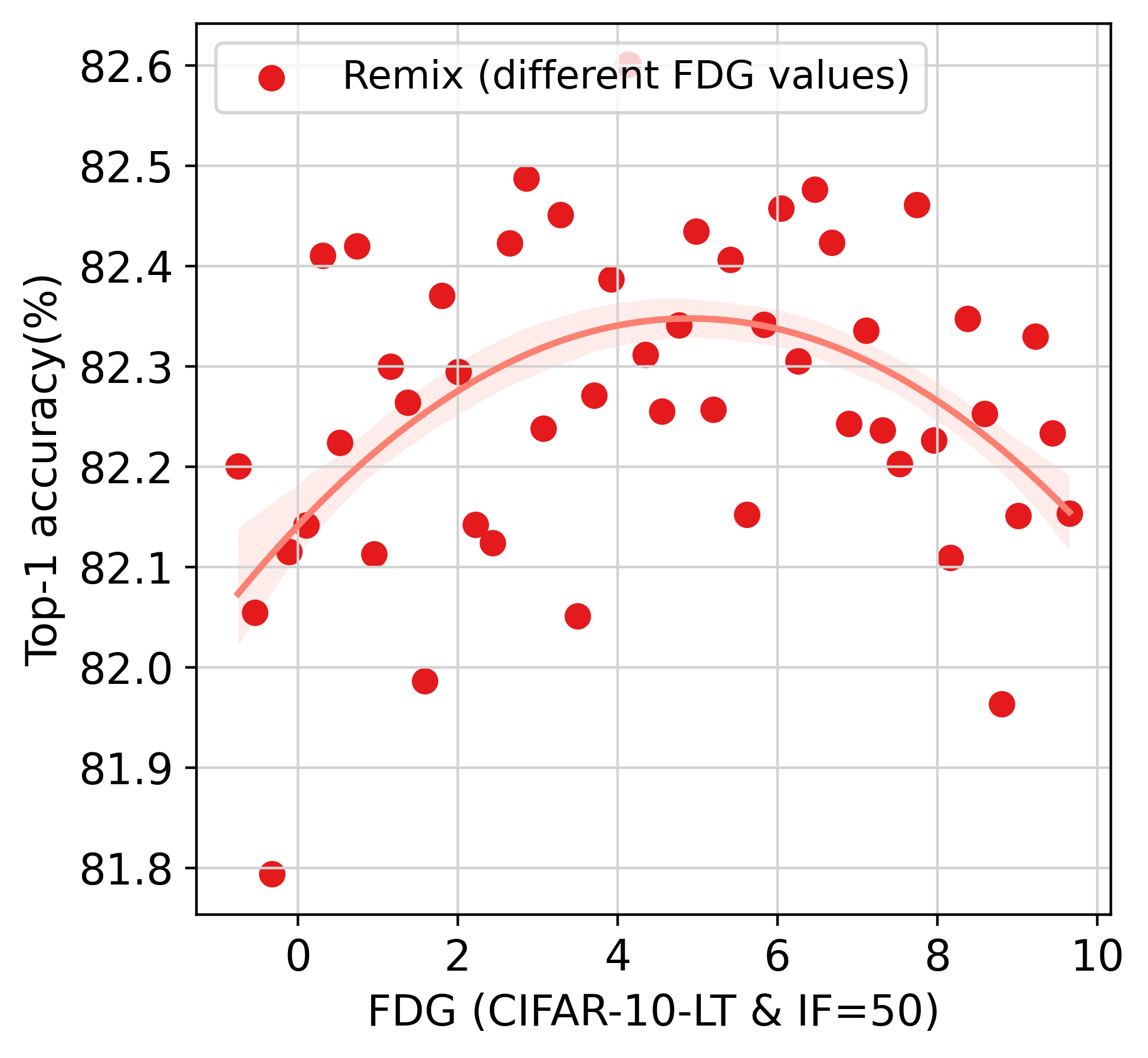}
	\end{minipage}
	\begin{minipage}{0.243\linewidth}
		\centering
		\includegraphics[width=1\linewidth]{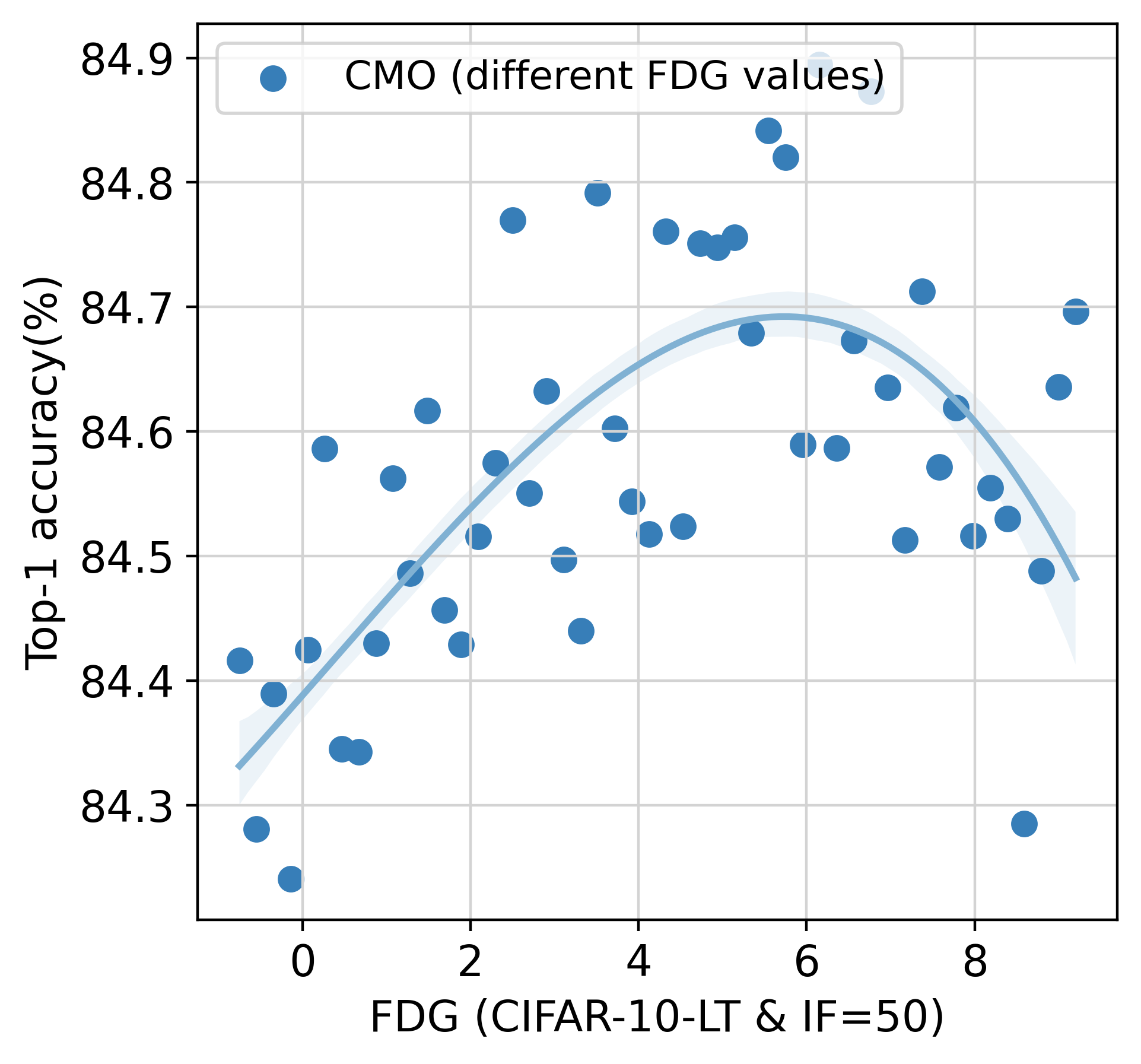}
	\end{minipage}
	\begin{minipage}{0.243\linewidth}
		\centering
		\includegraphics[width=1\linewidth]{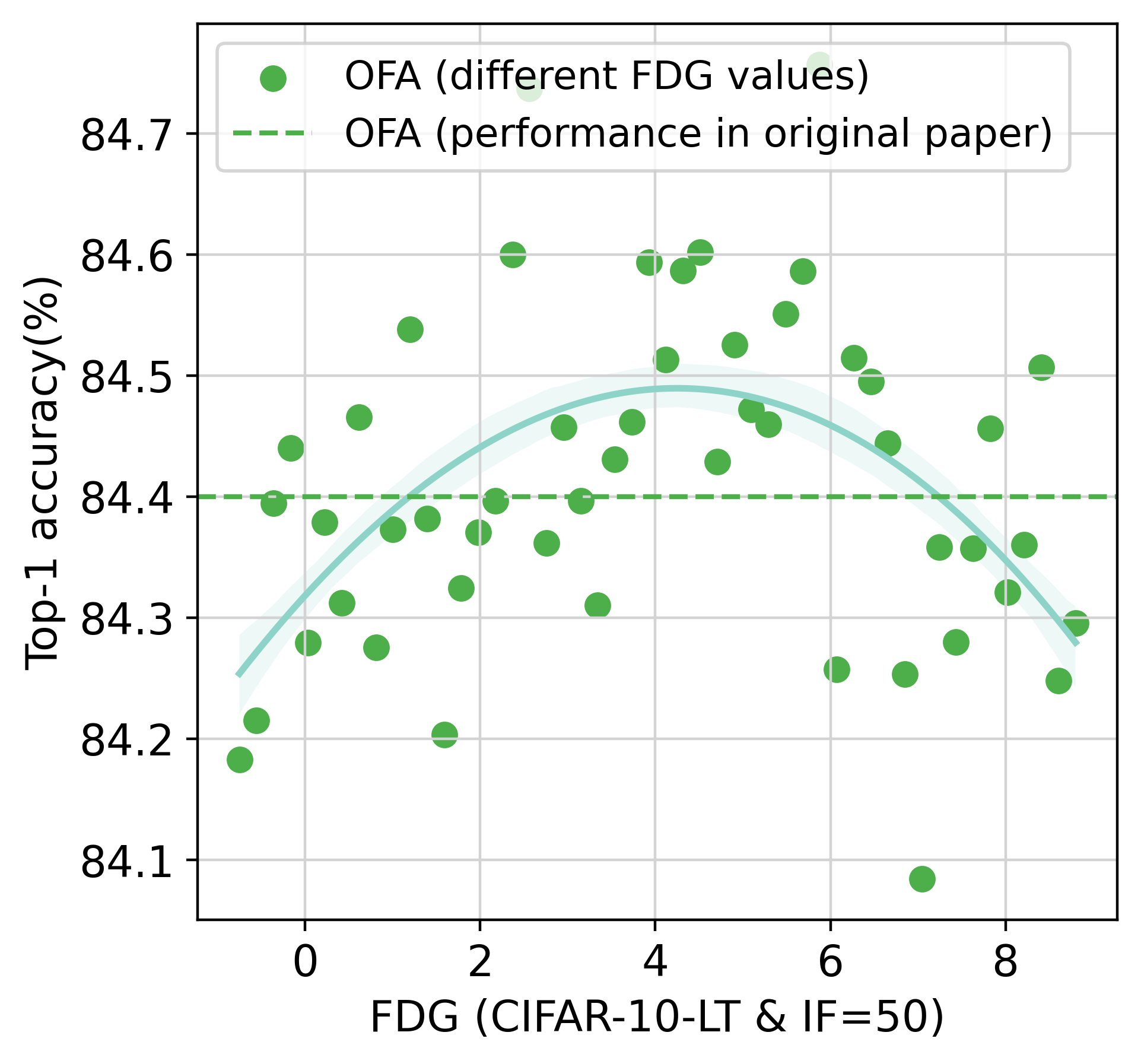}
	\end{minipage}
	\begin{minipage}{0.243\linewidth}
		\centering
		\includegraphics[width=1\linewidth]{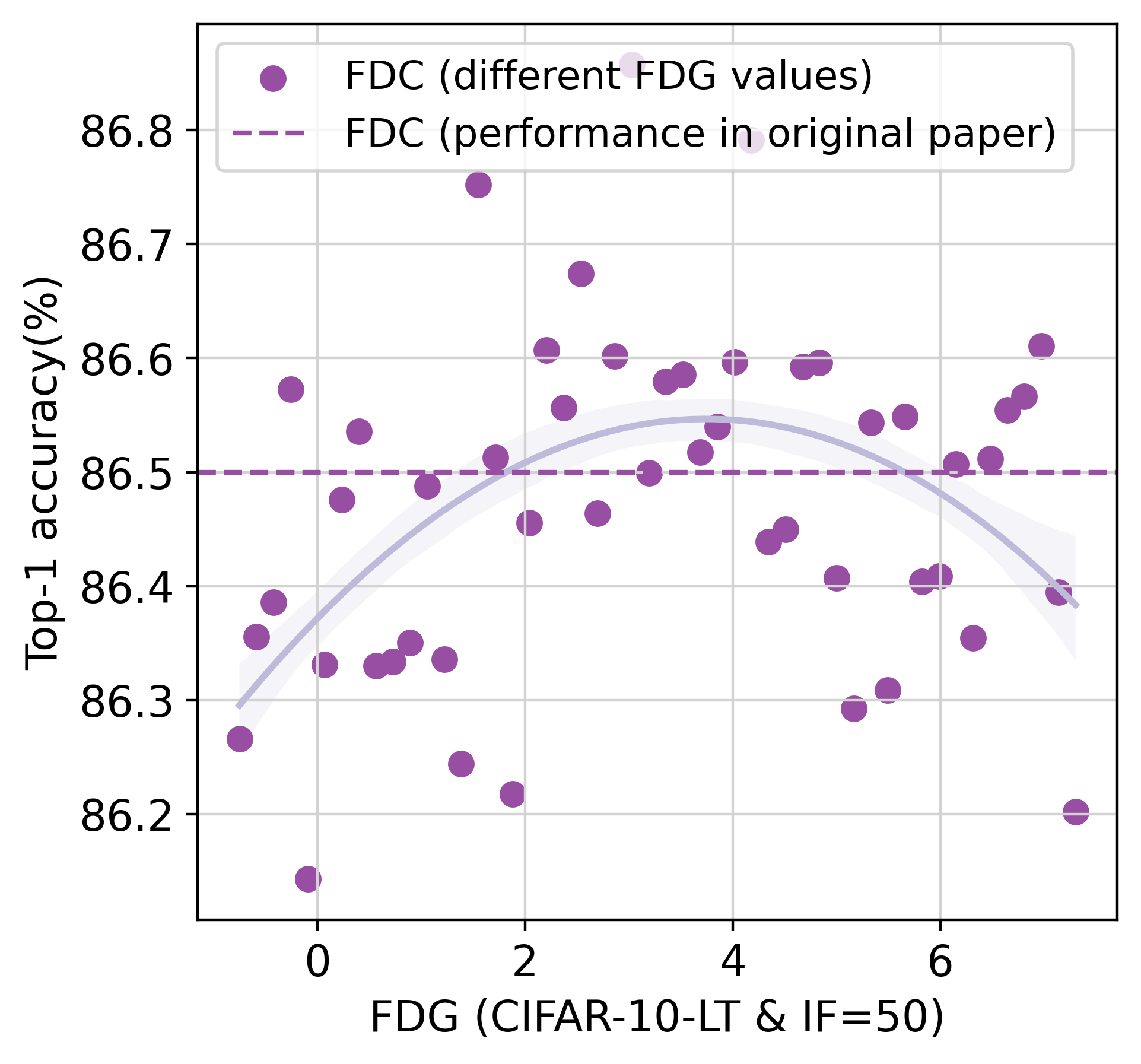}
	\end{minipage}

     \centering
	\begin{minipage}{0.243\linewidth}
		\centering
		\includegraphics[width=1\linewidth]{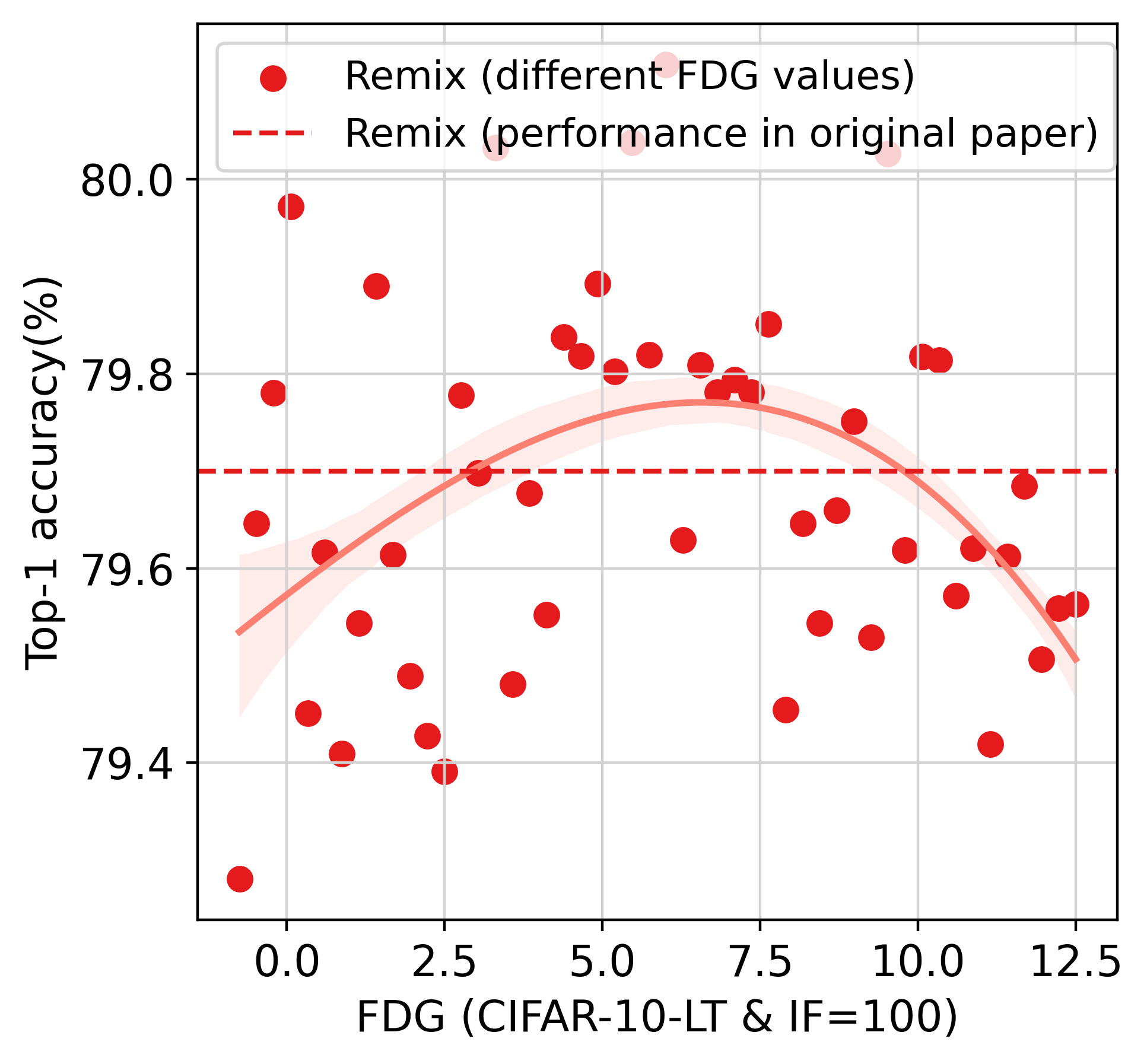}
	\end{minipage}
	\begin{minipage}{0.243\linewidth}
		\centering
		\includegraphics[width=1\linewidth]{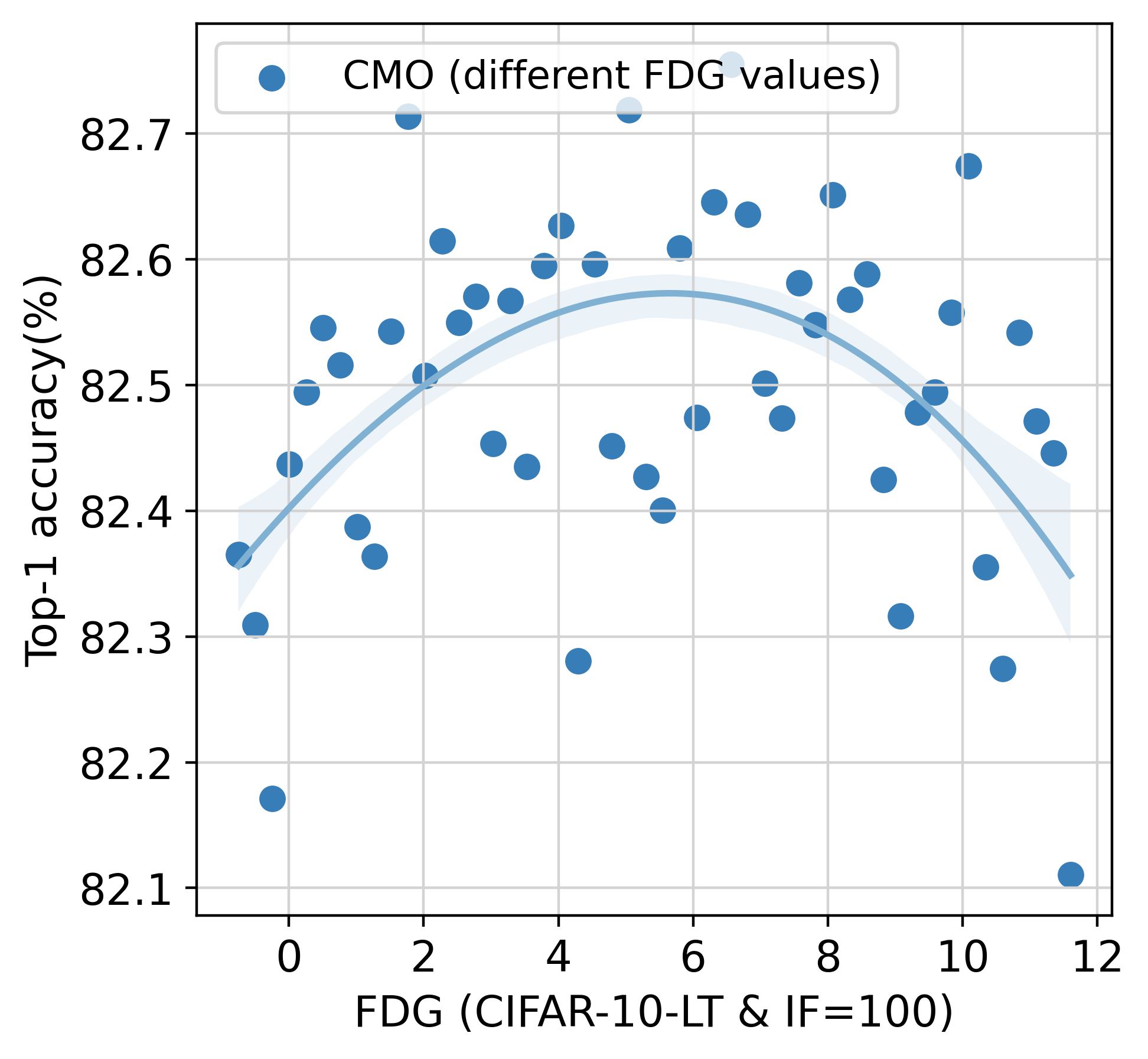}
	\end{minipage}
	\begin{minipage}{0.243\linewidth}
		\centering
		\includegraphics[width=1\linewidth]{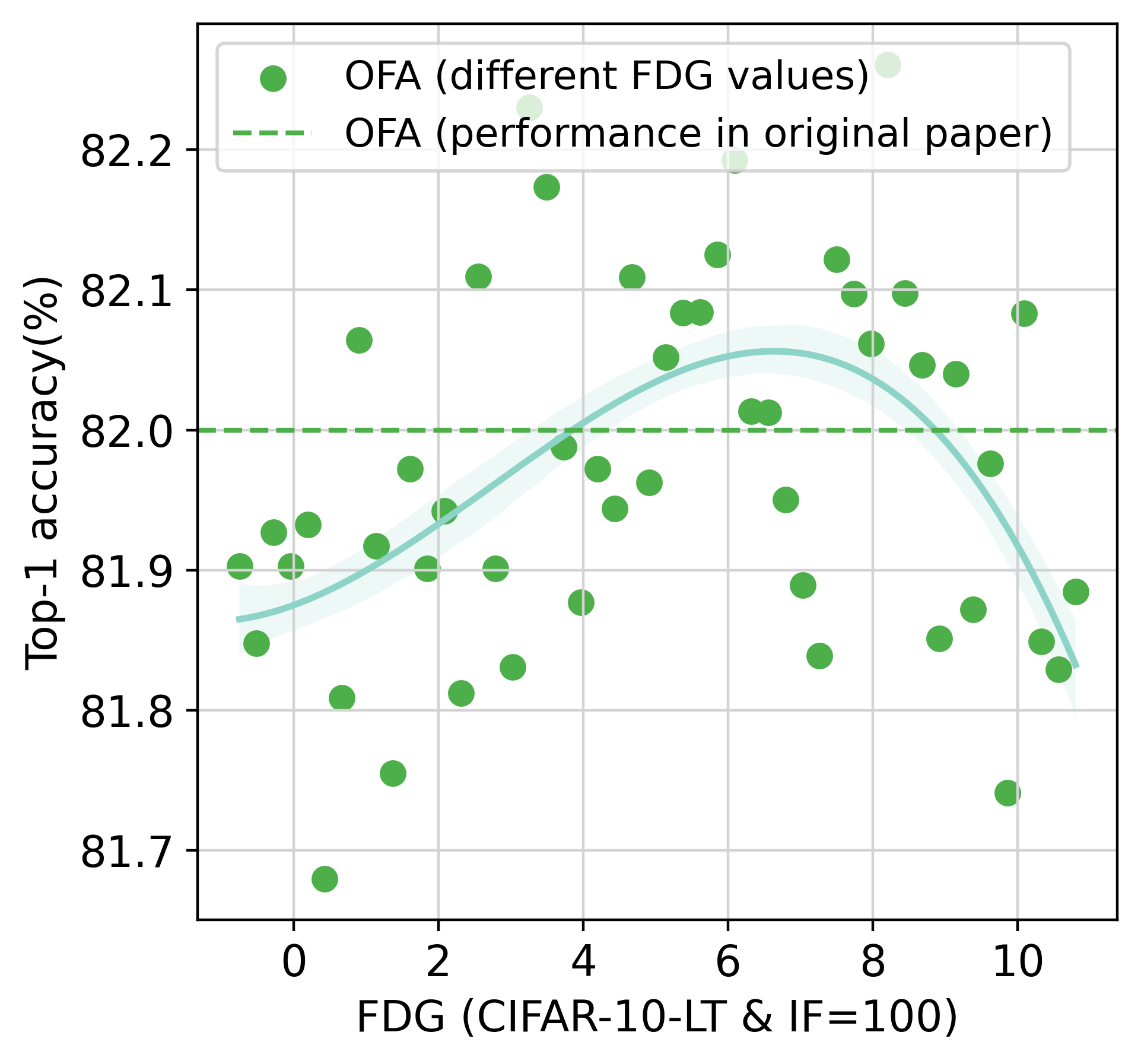}
	\end{minipage}
	\begin{minipage}{0.243\linewidth}
		\centering
		\includegraphics[width=1\linewidth]{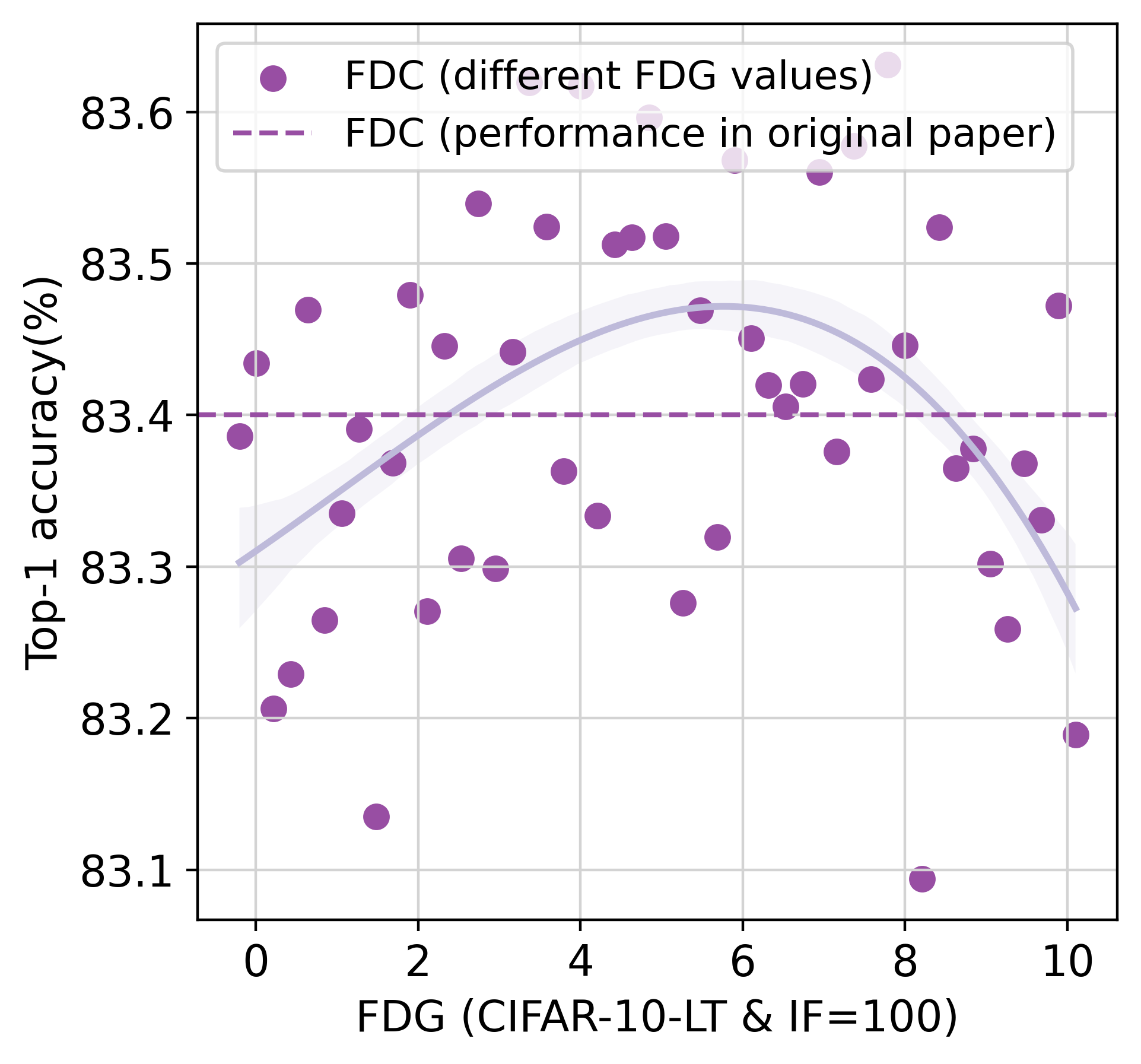}
	\end{minipage}
\caption{The performance of information augmentation with different levels of FDG on CIFAR-10-LT.}
\label{fig3}
\vskip -0.05in
\end{figure*}

\subsection{Appropriate FDG promotes long-tailed recognition}
\label{sec4.3}

The Imbalance Factor (IF) represents the degree of imbalance in a long-tail dataset, defined as the ratio between the number of training samples in the most common class and the class with the lowest occurrence frequency. We chose to conduct classification experiments using four information augmentation methods on six different datasets: CIFAR-10-LT (IF = $10, 50, 100$) and CIFAR-100-LT (IF = $10, 50, 100$) \cite{paper8}. Each of the four methods generated $50$ sets of effective augmented data for each dataset, resulting in a total of $1200$ classification experiments. Following established settings, the backbone network used was ResNet-32, and the experimental parameters for all four methods remained consistent with the original papers.

\begin{figure*}[!t]
\centering
	\begin{minipage}{0.243\linewidth}
		\centering
		\includegraphics[width=1\linewidth]{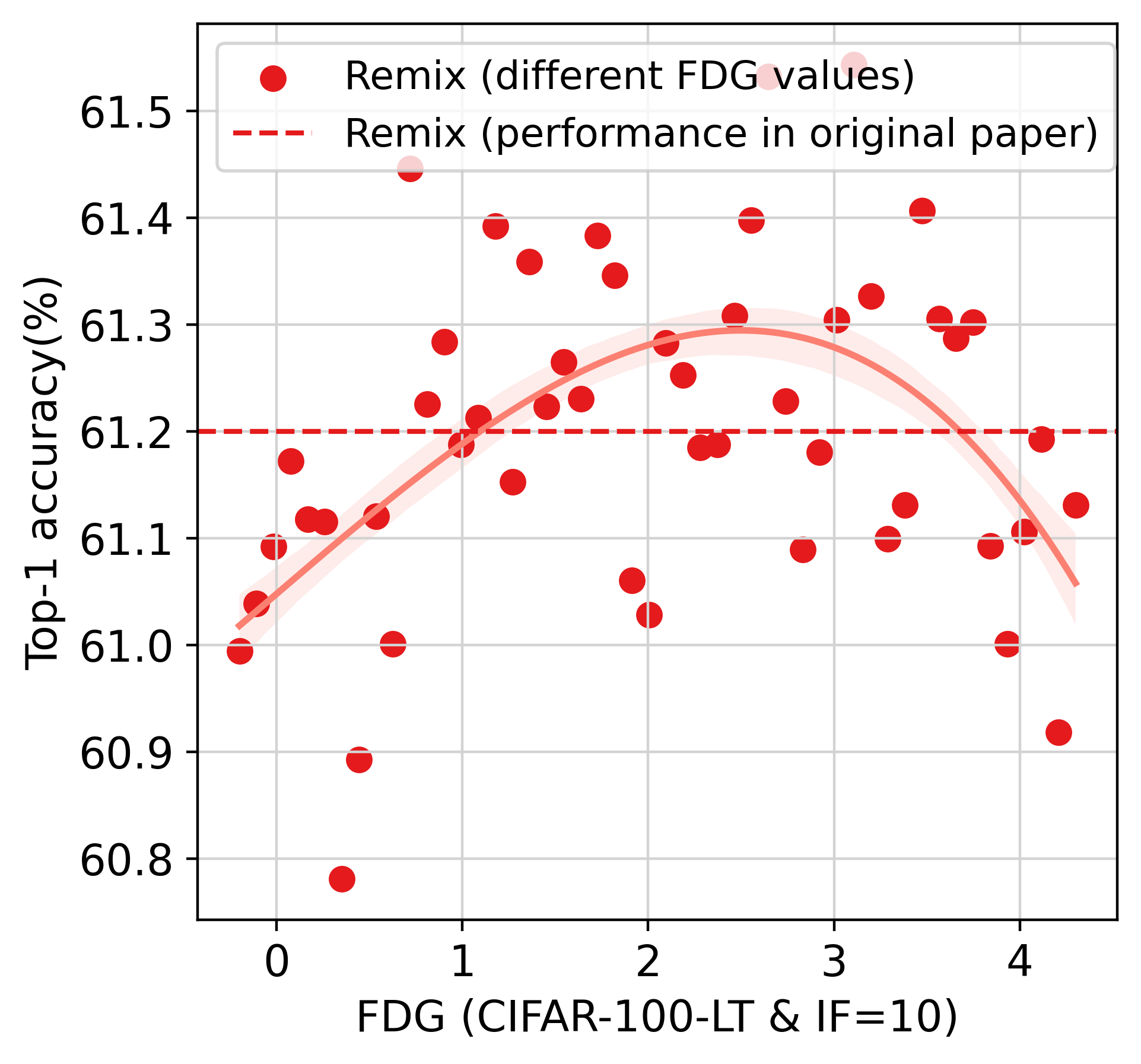}
	\end{minipage}
	\begin{minipage}{0.243\linewidth}
		\centering
		\includegraphics[width=1\linewidth]{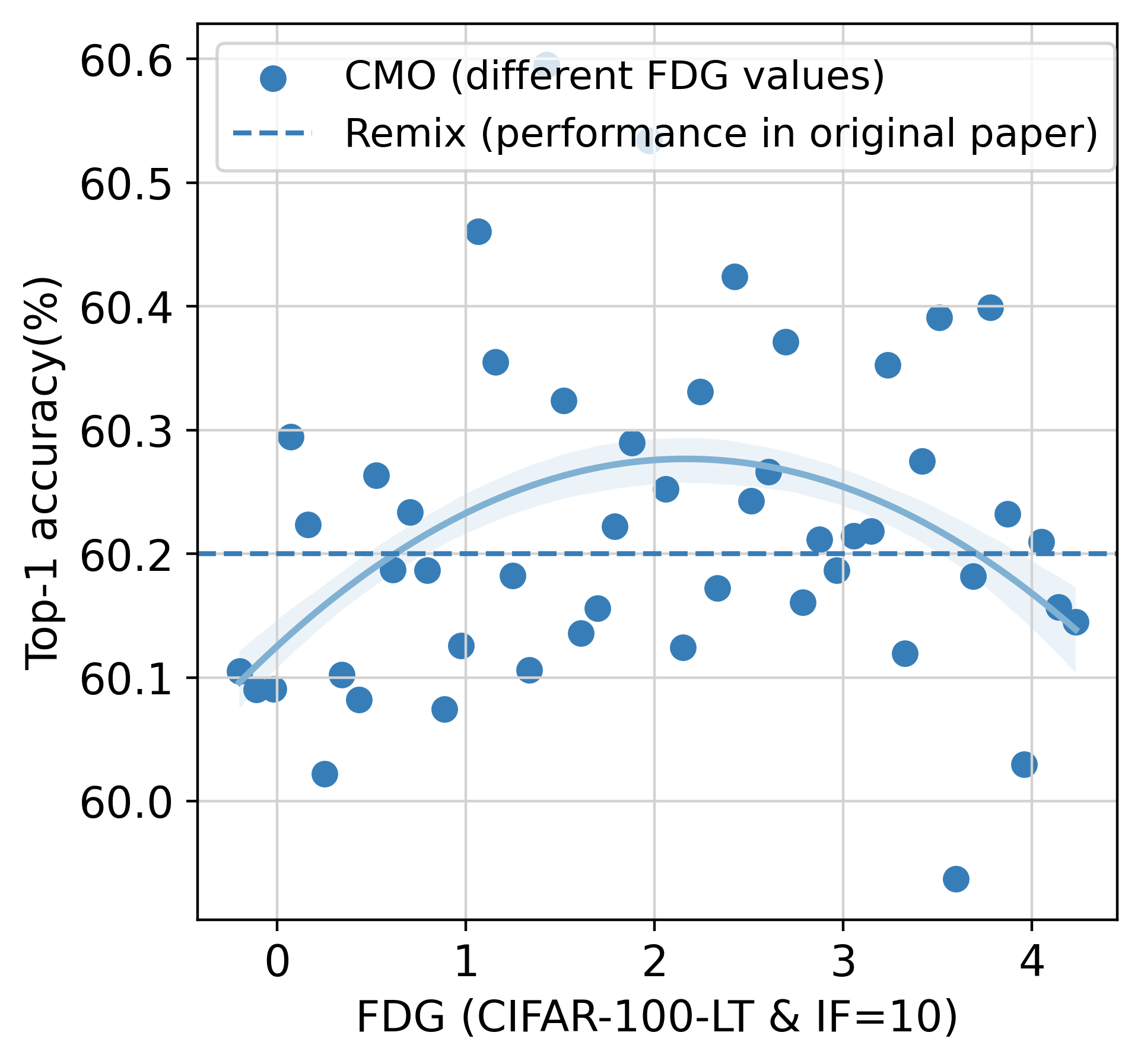}
	\end{minipage}
	\begin{minipage}{0.243\linewidth}
		\centering
		\includegraphics[width=1\linewidth]{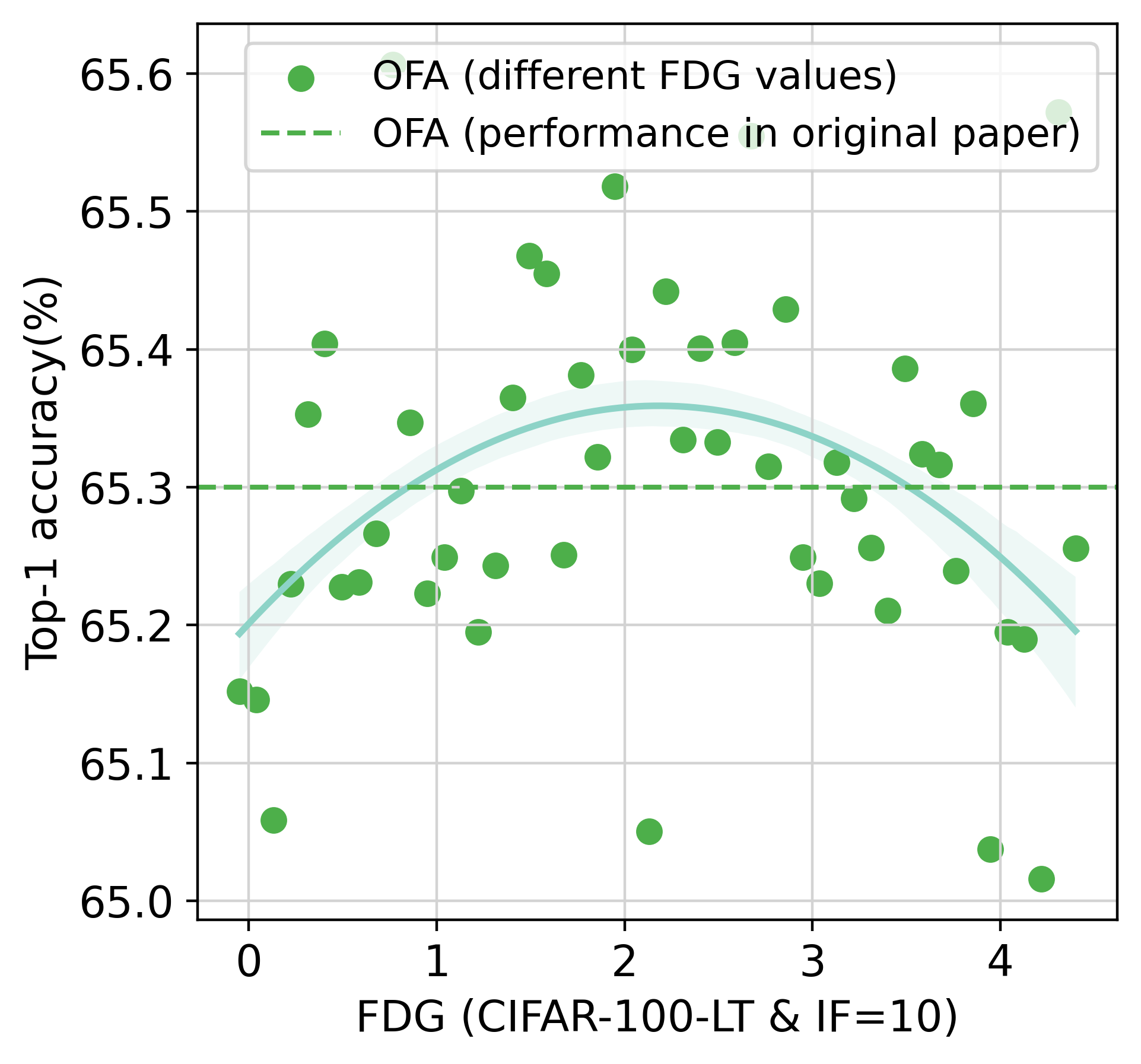}
	\end{minipage}
	\begin{minipage}{0.243\linewidth}
		\centering
		\includegraphics[width=1\linewidth]{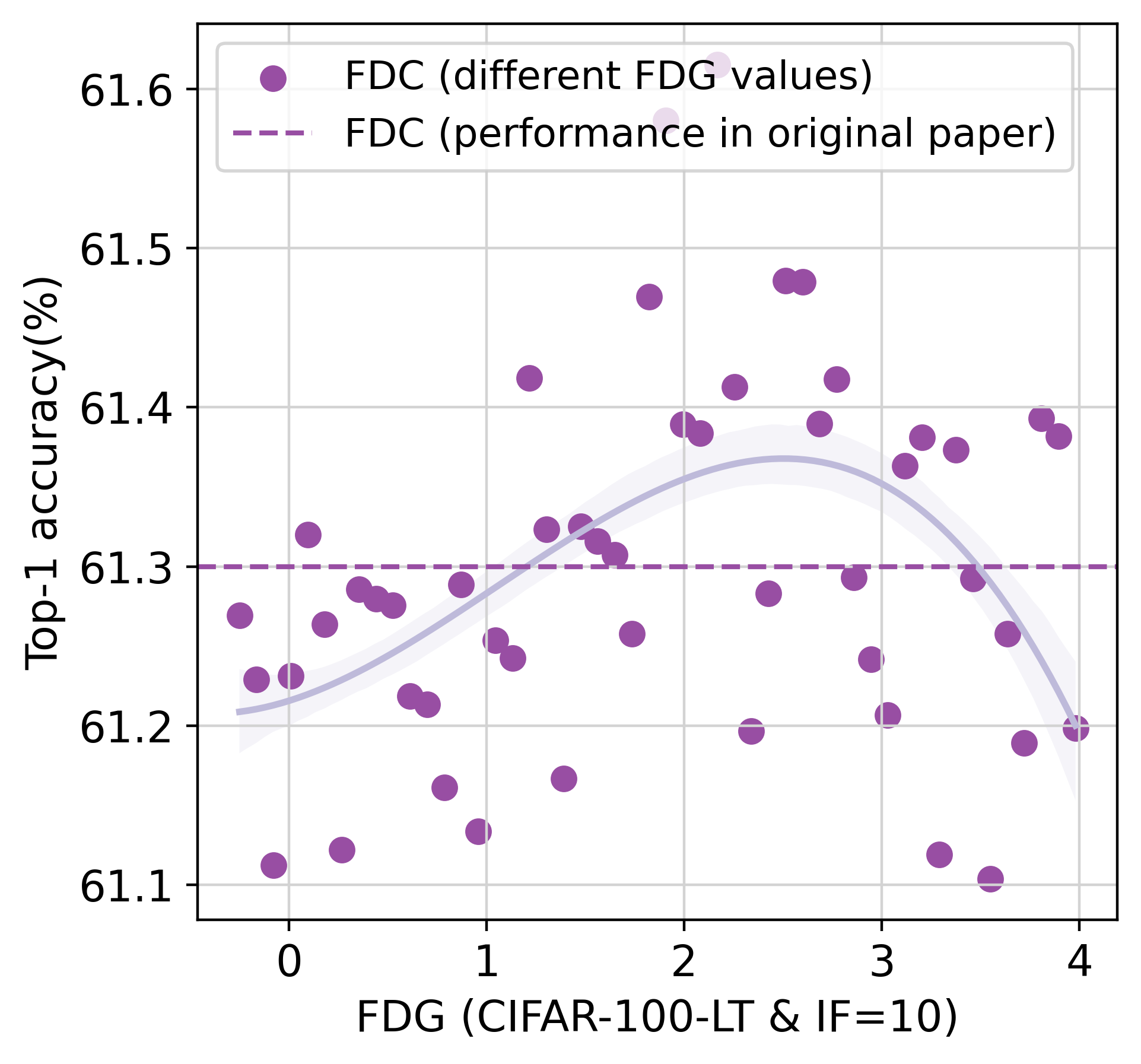}
	\end{minipage} 

     \centering
	\begin{minipage}{0.243\linewidth}
		\centering
		\includegraphics[width=1\linewidth]{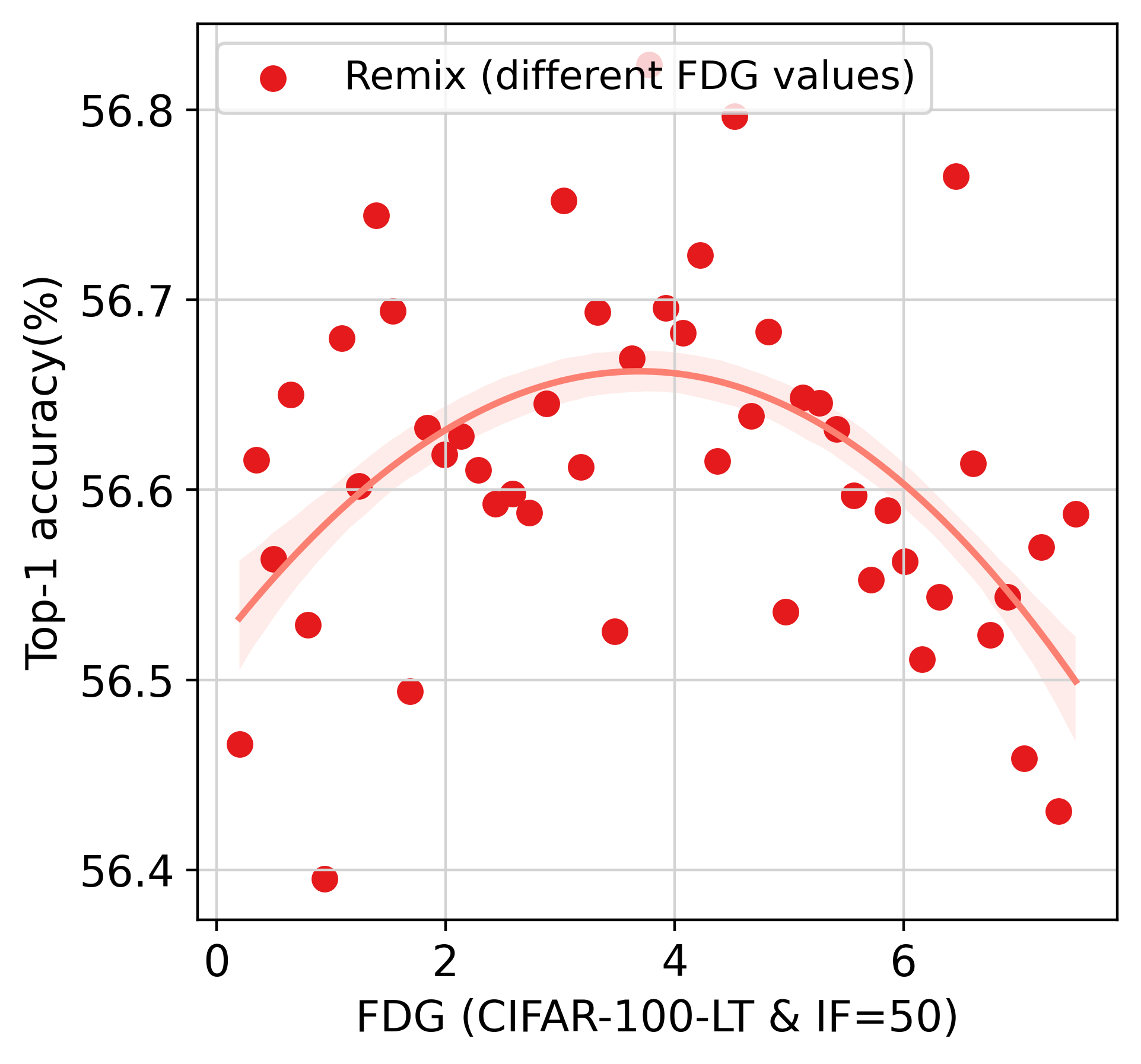}
	\end{minipage}
	\begin{minipage}{0.243\linewidth}
		\centering
		\includegraphics[width=1\linewidth]{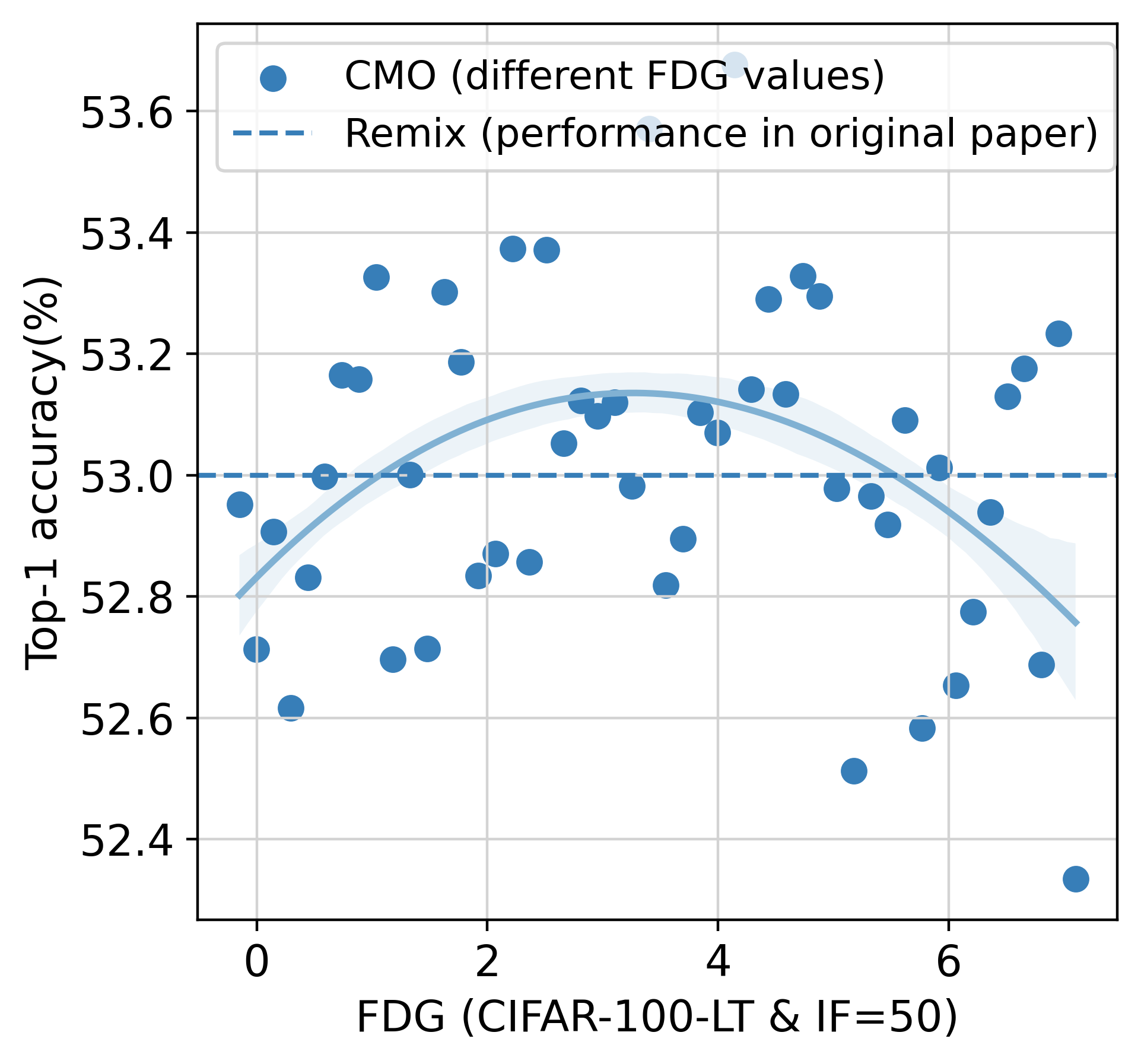}
	\end{minipage}
	\begin{minipage}{0.243\linewidth}
		\centering
		\includegraphics[width=1\linewidth]{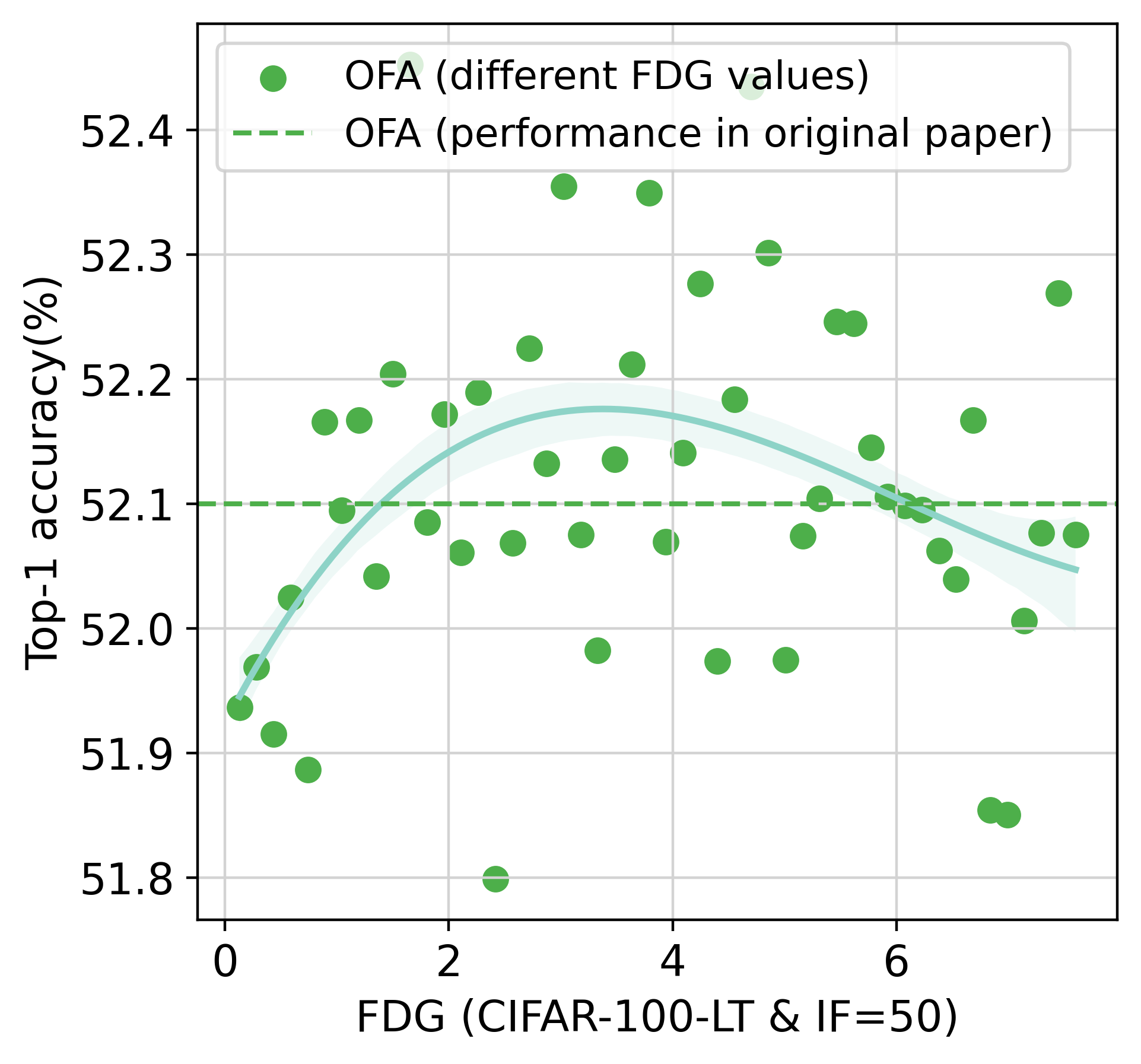}
	\end{minipage}
	\begin{minipage}{0.243\linewidth}
		\centering
		\includegraphics[width=1\linewidth]{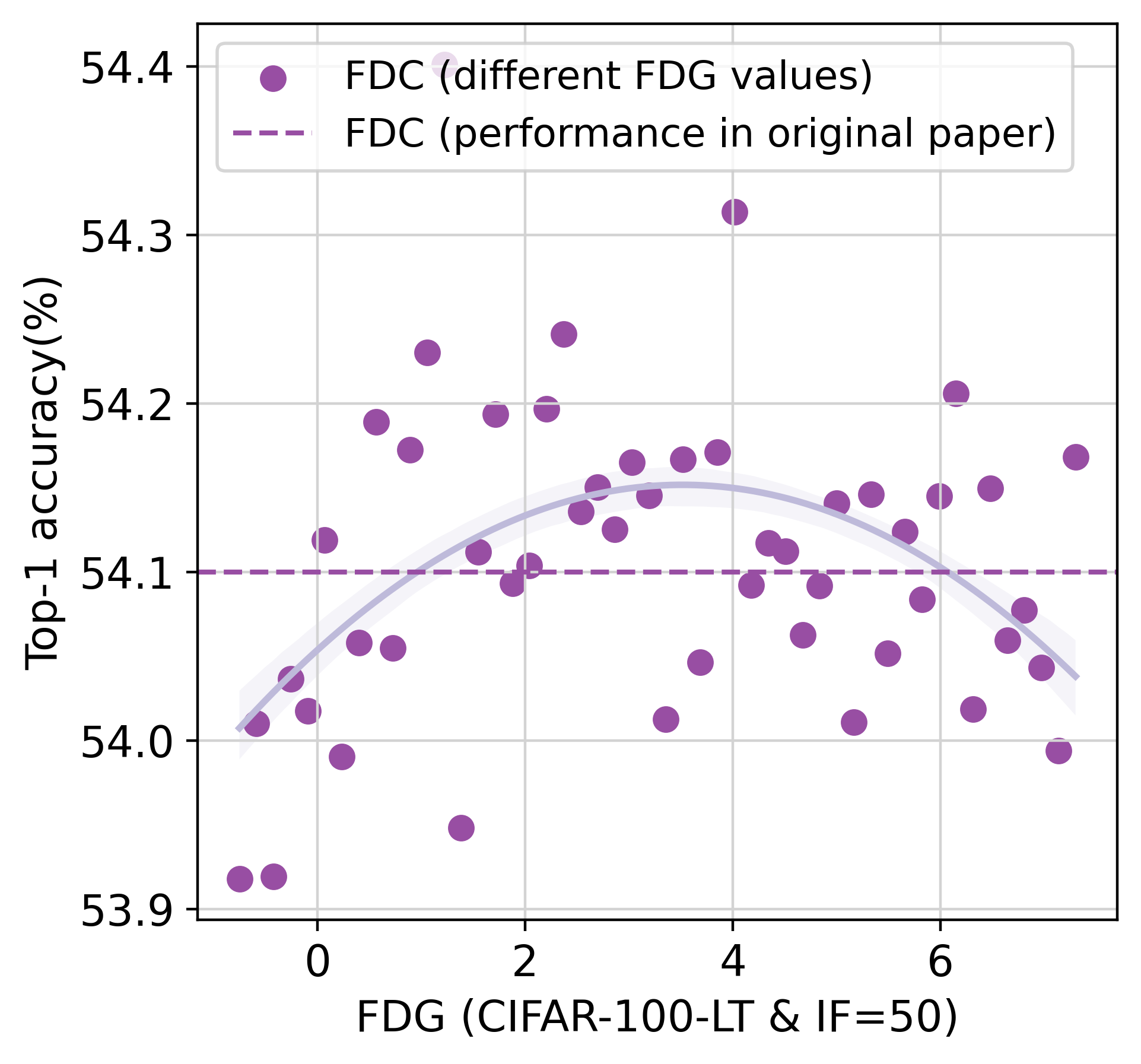}
	\end{minipage}

     \centering
	\begin{minipage}{0.243\linewidth}
		\centering
		\includegraphics[width=1\linewidth]{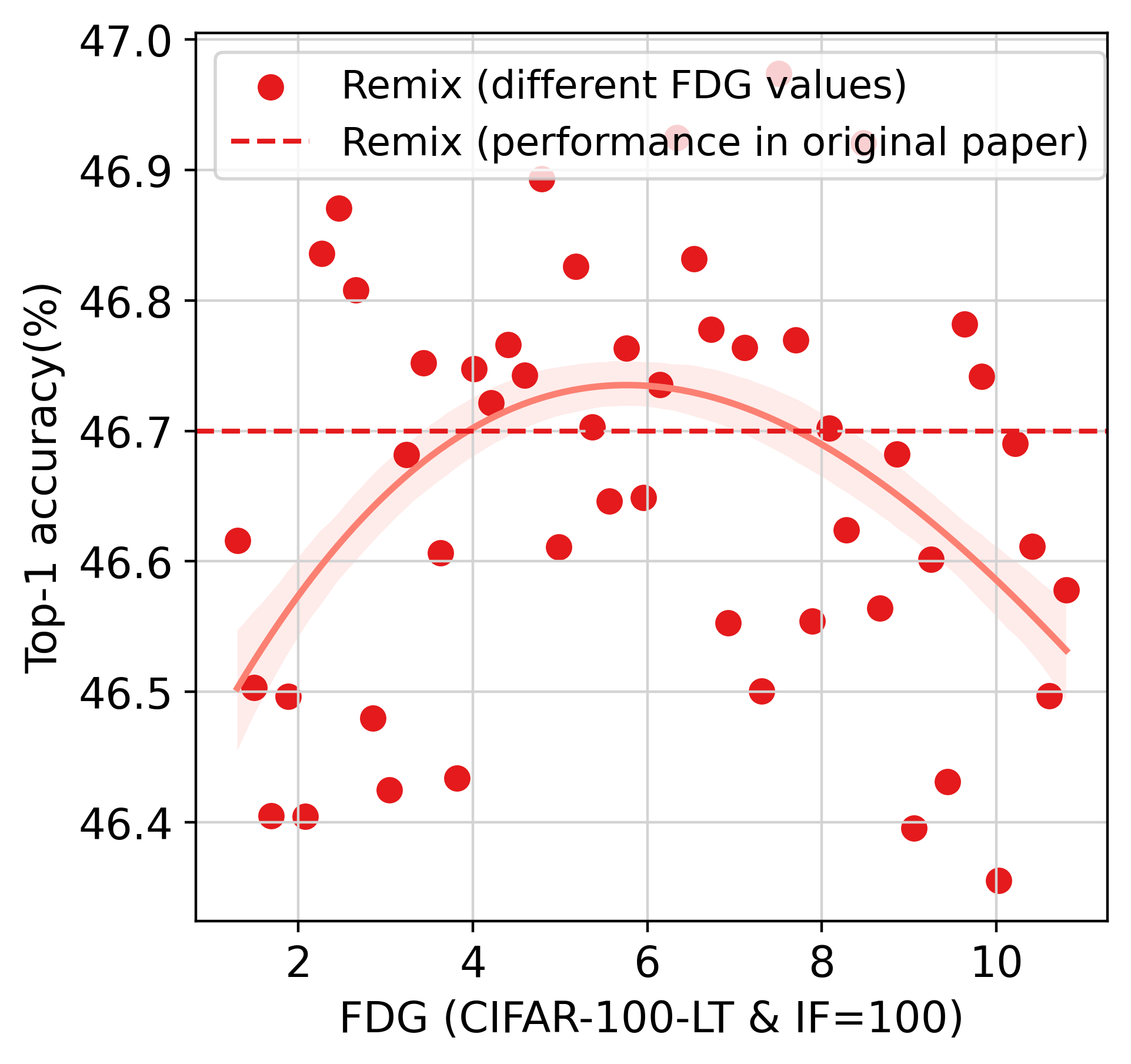}
	\end{minipage}
	\begin{minipage}{0.243\linewidth}
		\centering
		\includegraphics[width=1\linewidth]{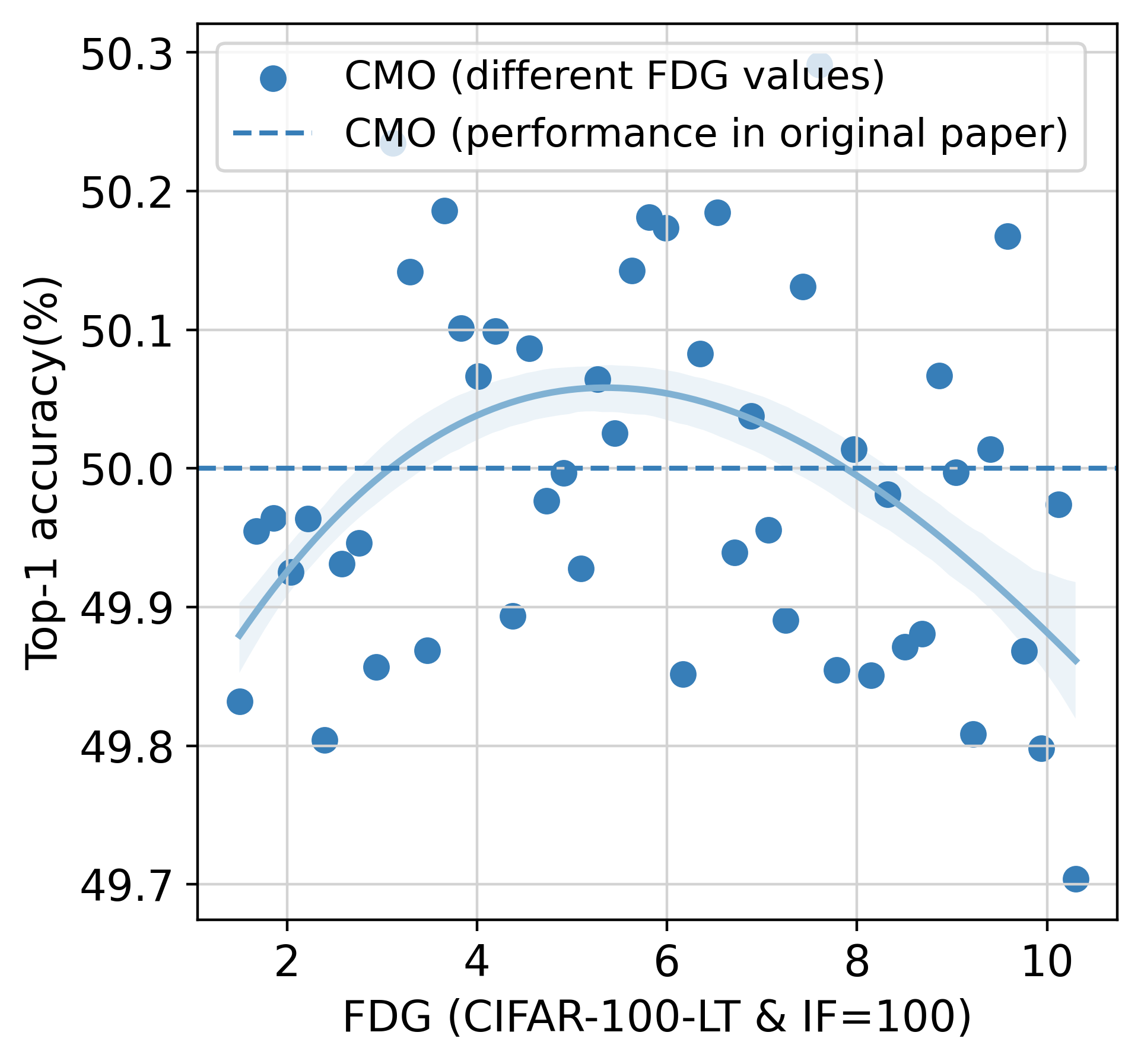}
	\end{minipage}
	\begin{minipage}{0.243\linewidth}
		\centering
		\includegraphics[width=1\linewidth]{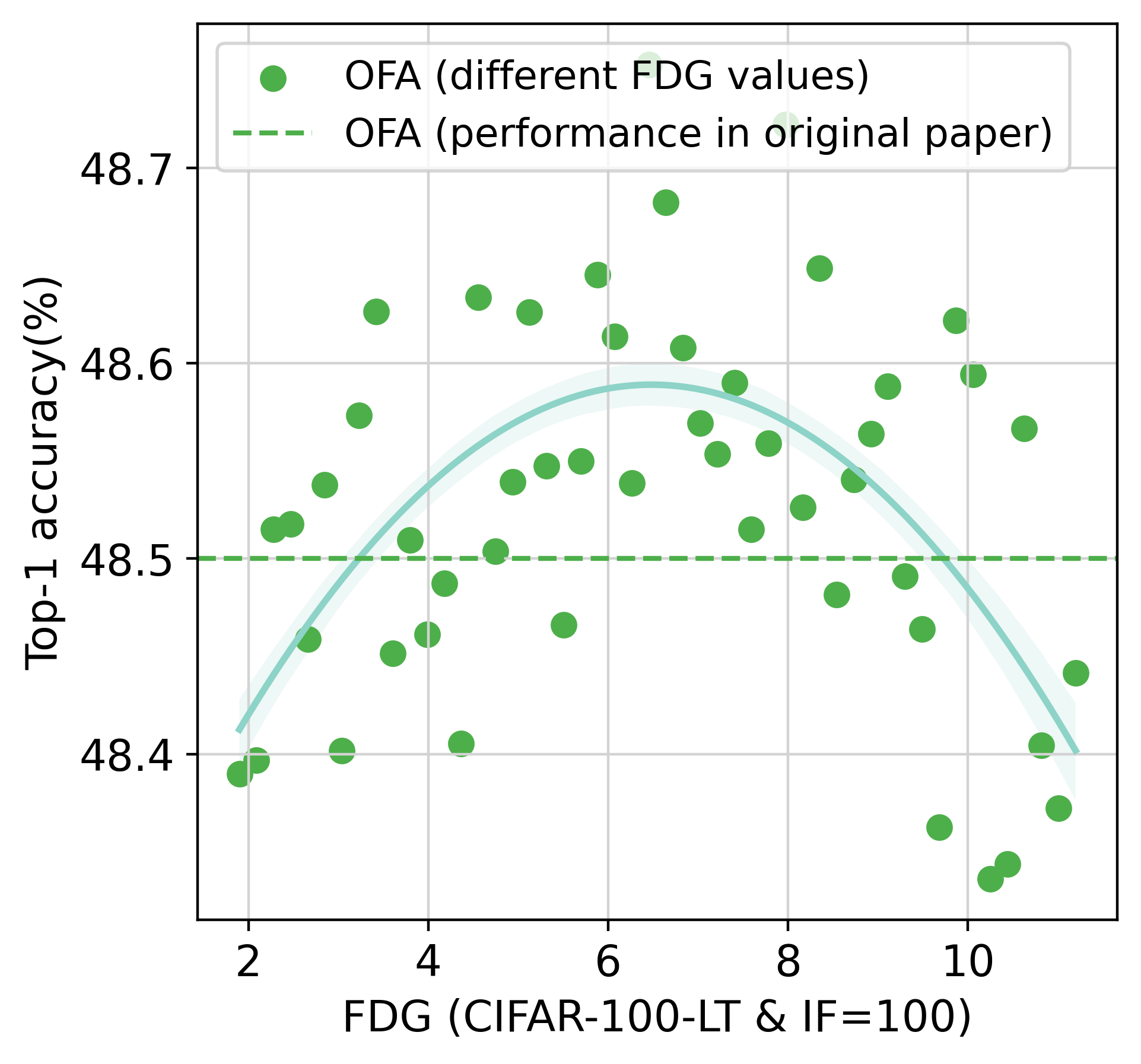}
	\end{minipage}
	\begin{minipage}{0.243\linewidth}
		\centering
		\includegraphics[width=1\linewidth]{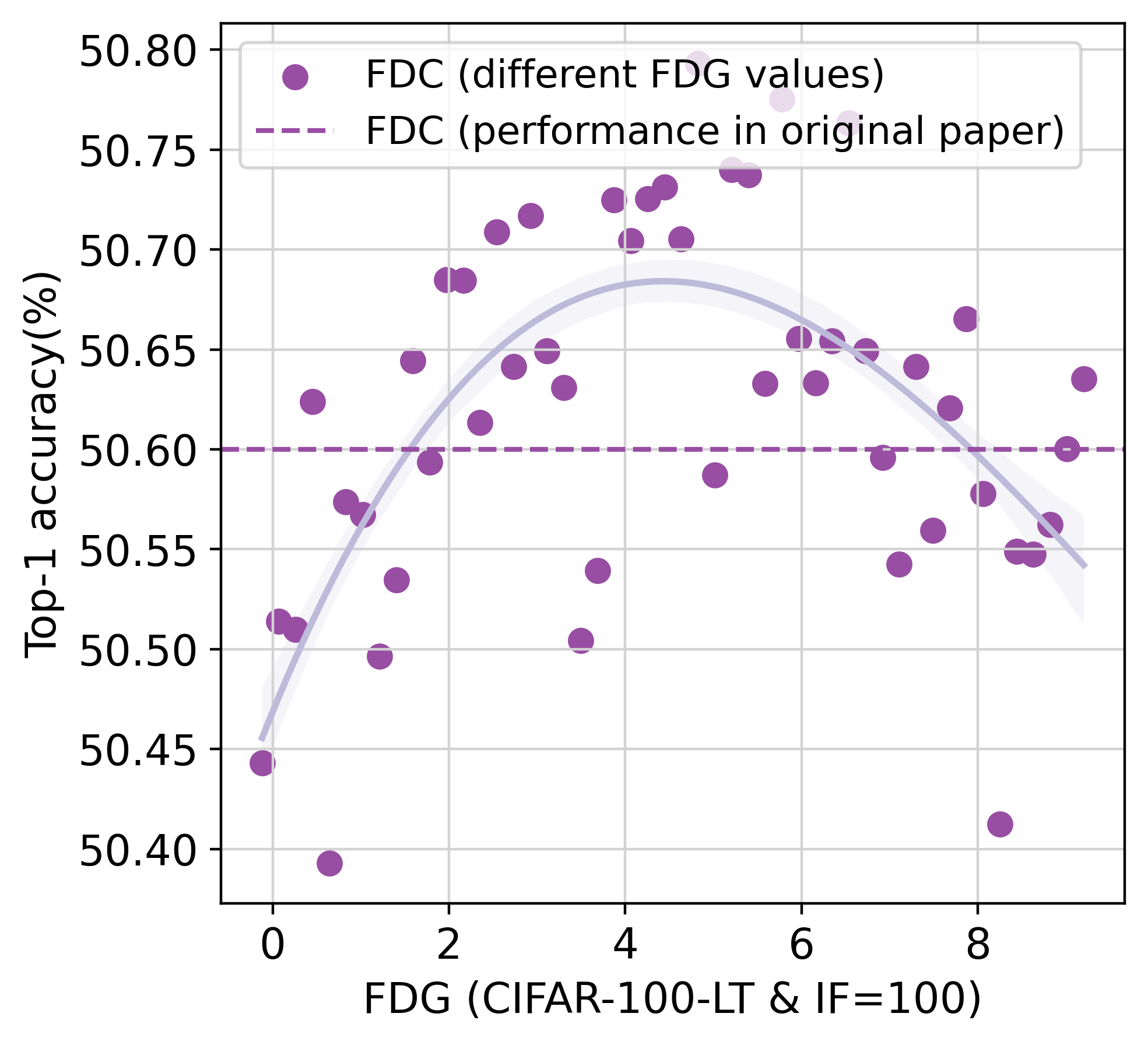}
	\end{minipage}
\vskip -0.05in
\caption{The performance of information augmentation with different levels of FDG on CIFAR-100-LT.}
\label{fig4}
\vskip -0.05in
\end{figure*}

The experimental results on CIFAR-10-LT are shown in Fig.\ref{fig3}. Overall, it can be observed that the performance of information augmentation methods is optimized when the FDG corresponding to augmented data falls within an appropriate range. Furthermore, as the imbalance degree of the dataset increases, information augmentation methods need to generate augmented data that can provide more feature diversity gain. The experimental results on CIFAR-100-LT (see Fig.\ref{fig4}) are generally consistent with those on CIFAR-10-LT. It's worth noting that the original CMO paper did not provide results on CIFAR-10-LT. Therefore, we only conducted classification experiments without providing a baseline for comparison. However, this does not hinder us from drawing the main experimental conclusions.

Overall, our experiments demonstrate that by guiding information augmentation methods to generate more appropriate samples, it is possible to further improve the performance of long-tailed models. In the field of long-tailed learning, data-centric approaches should be given more attention and should work in conjunction with model-centric methods to foster the development of this field. In the next chapter, we will delve into the fundamental tasks of data-centric long-tailed learning, providing more research insights for future studies.

\section{Data-Centric Long-Tailed Learning}
\label{sec5}

\begin{figure*}[t]
\centering
\includegraphics[width=7.22in]{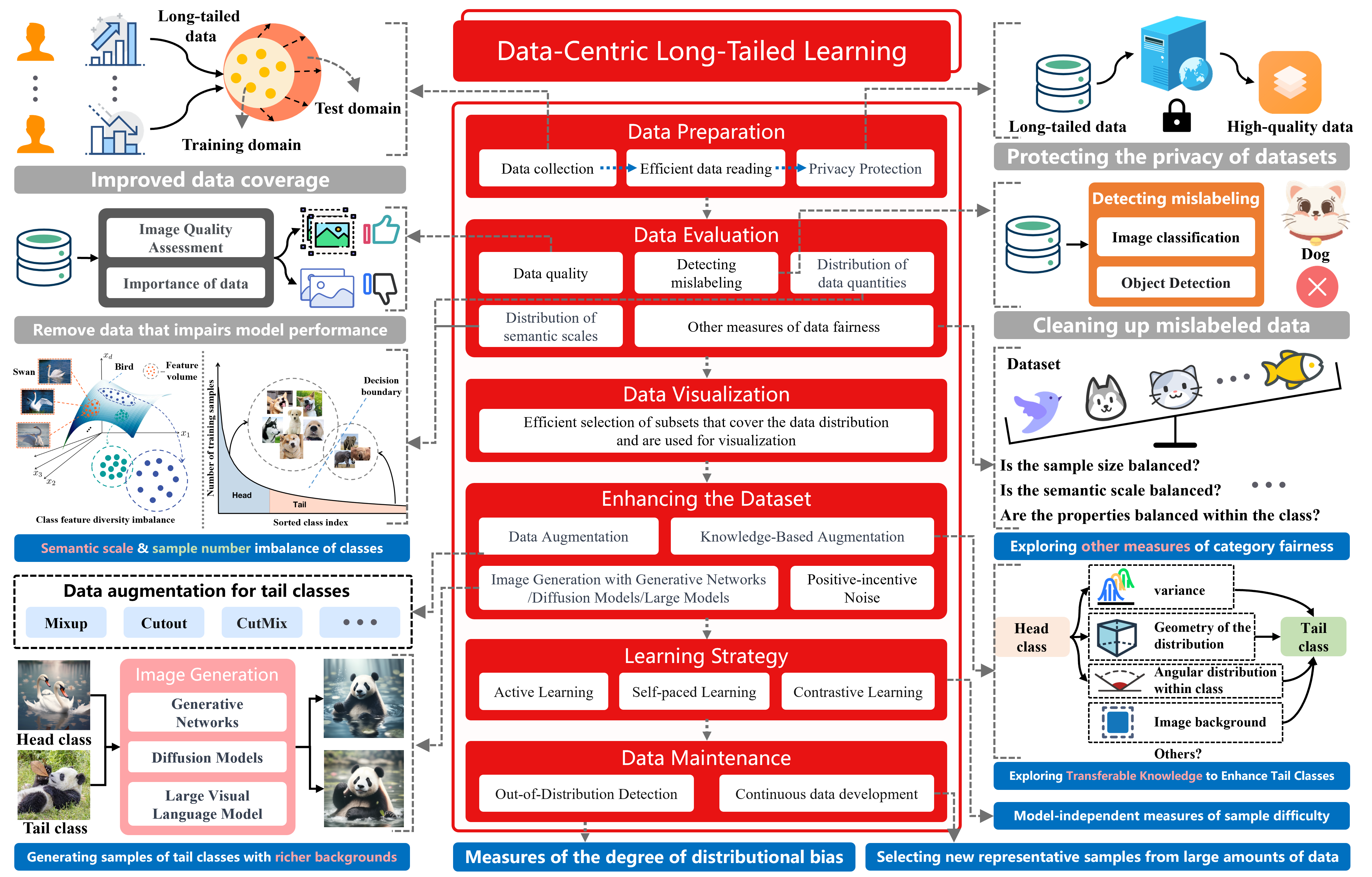}
\caption{The core components and fundamental tasks of the data-centric long-tailed learning framework. The central red box represents the system's core components, with the content on both sides corresponding to the fundamental tasks of these core components. Developing and advancing these fundamental tasks will enhance the performance of the core components, thereby optimizing the system.}
\label{fig5}
\end{figure*}

As algorithms continue to rapidly evolve, the diminishing returns from improving model structures become increasingly evident. For instance, the rapid iterations achieved by ChatGPT are primarily attributed to more rigorous data engineering. In specific application scenarios, data often exhibits a long-tailed distribution, making it challenging to obtain samples from rare categories. When training samples fail to adequately cover the situations encountered in practice, relying solely on improvements to the model structure is insufficient to address the biases in artificial intelligence systems \cite{paper67,paper63,paper72}. The most fundamental solution is to enrich the training dataset with more diverse and abundant samples, which necessitates additional knowledge for introducing or generating augmented samples. Acquiring high-quality data has become a core requirement in the current AI development landscape \cite{paper66}. Based on this, we propose a data-centric long-tailed learning framework. As shown in Fig.\ref{fig5}, by developing the components and fundamental tasks within this framework, we aim to provide users with an end-to-end data service. Specifically, when users collect a set of low-quality data that covers a limited and imbalanced range, they can input these data into the data-centric long-tailed learning system. This system will \textbf{provide users with a set of higher-quality data for training purposes}. Below, we will provide a detailed introduction to the core components and fundamental tasks depicted in Fig.\ref{fig5}.

\begin{figure}[t]
\centering
\includegraphics[width=3.5in]{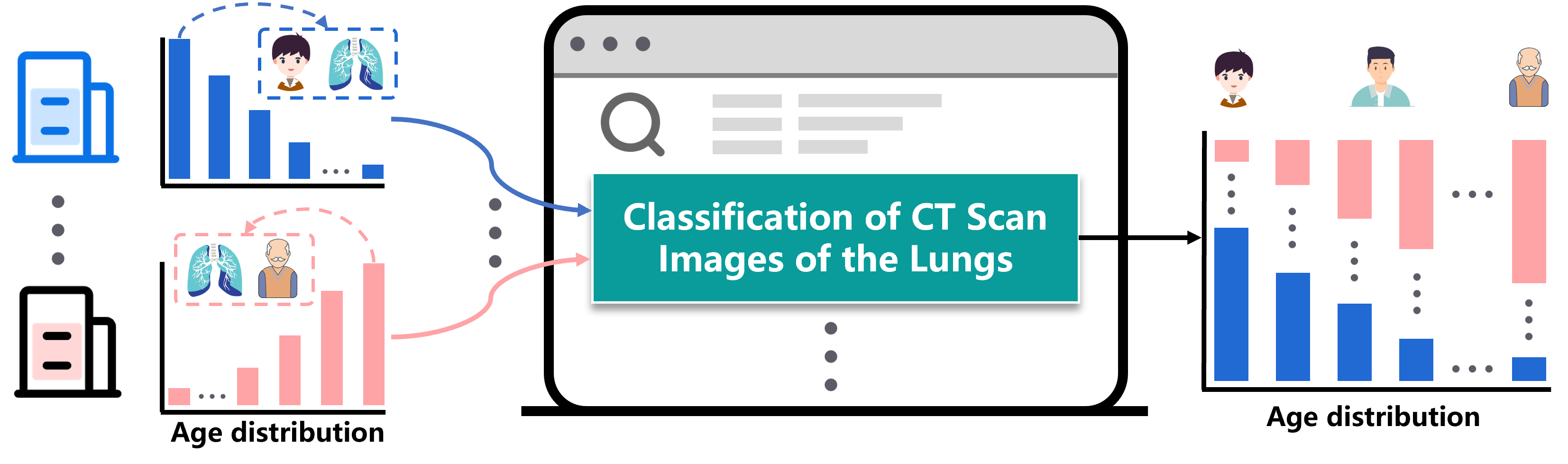}
\vskip -0.05in
\caption{An open-source platform for collaborative dataset creation. It selects high-quality data from various sources while maintaining an unbiased dataset by analyzing sample distributions.}
\label{fig6}
\end{figure}

\subsection{Data Preparation}
\label{sec5.1}

\subsubsection{Data collection}

Before developing an artificial intelligence system with specific functionalities, it is necessary to determine the categories of required data and estimate the number of samples for each category. When the distribution of collected data effectively covers the true distribution of all categories, this training dataset can be considered of high quality \cite{paper66, paper5}. However, in some specialized scenarios, it is often challenging to collect samples for certain categories \cite{paper97,paper98,paper99,paper100}. For example, in medical imaging used for disease diagnosis, the majority of images are of normal conditions, with images showing evident abnormalities representing only a small fraction \cite{paper95}. This scarcity is particularly significant in the case of rare diseases \cite{paper96}.

To ensure that the collected sample distribution covers the situations that may be encountered in practice as comprehensively as possible, it is beneficial to obtain data from multiple sources \cite{paper101}. For instance, samples collected from a children's hospital and a general hospital may differ in age, and to prevent the model from biasing towards a particular age group, both parties can collaborate to create a shared dataset. As shown in Fig.\ref{fig6}, establishing an open-source platform for collaborative dataset creation is a feasible solution. This open-source platform can be partitioned according to different task types and data types, thereby enhancing the quality of datasets in various specialized scenarios. \textbf{It should be noted that in real-world applications, there may be long-tailed distributions across multiple attributes}. Therefore, when creating datasets, it is important to balance each attribute as much as possible.

Furthermore, it is essential to consider that some users are unwilling to share data of high value. Therefore, it is necessary to propose a \textbf{federated learning framework for information augmentation and data expansion}, which we will discuss in the following subsection.

\subsubsection{Efficient Data Retrieval and Privacy Protection}

When a dataset collected by users exhibits biases or other defects, how efficiently it can be accessed and enhanced is crucial for the system's ability to be widely deployed. As data becomes higher-dimensional or involves parallel processing for multiple tasks, the computational demands can significantly increase. Therefore, designing efficient data retrieval and processing mechanisms is a critical aspect.
Additionally, valuable data in sensitive domains needs to be protected. When users want to optimize their dataset without exposing it to data processors, privacy-preserving computation can play a significant role \cite{paper102,paper103,paper105}. One possible approach is for users to provide the server used for data computation and storage, while the service provider encrypts and protects the algorithms.
In addition to the need to protect privacy between users and service providers, there may be situations where users are unwilling to share data with other users \cite{paper104,paper106}. Therefore, it is crucial to explore methods for augmenting the dataset of all users without disclosing individual user data.

\begin{figure}[h]
\centering
\includegraphics[width=3.5in]{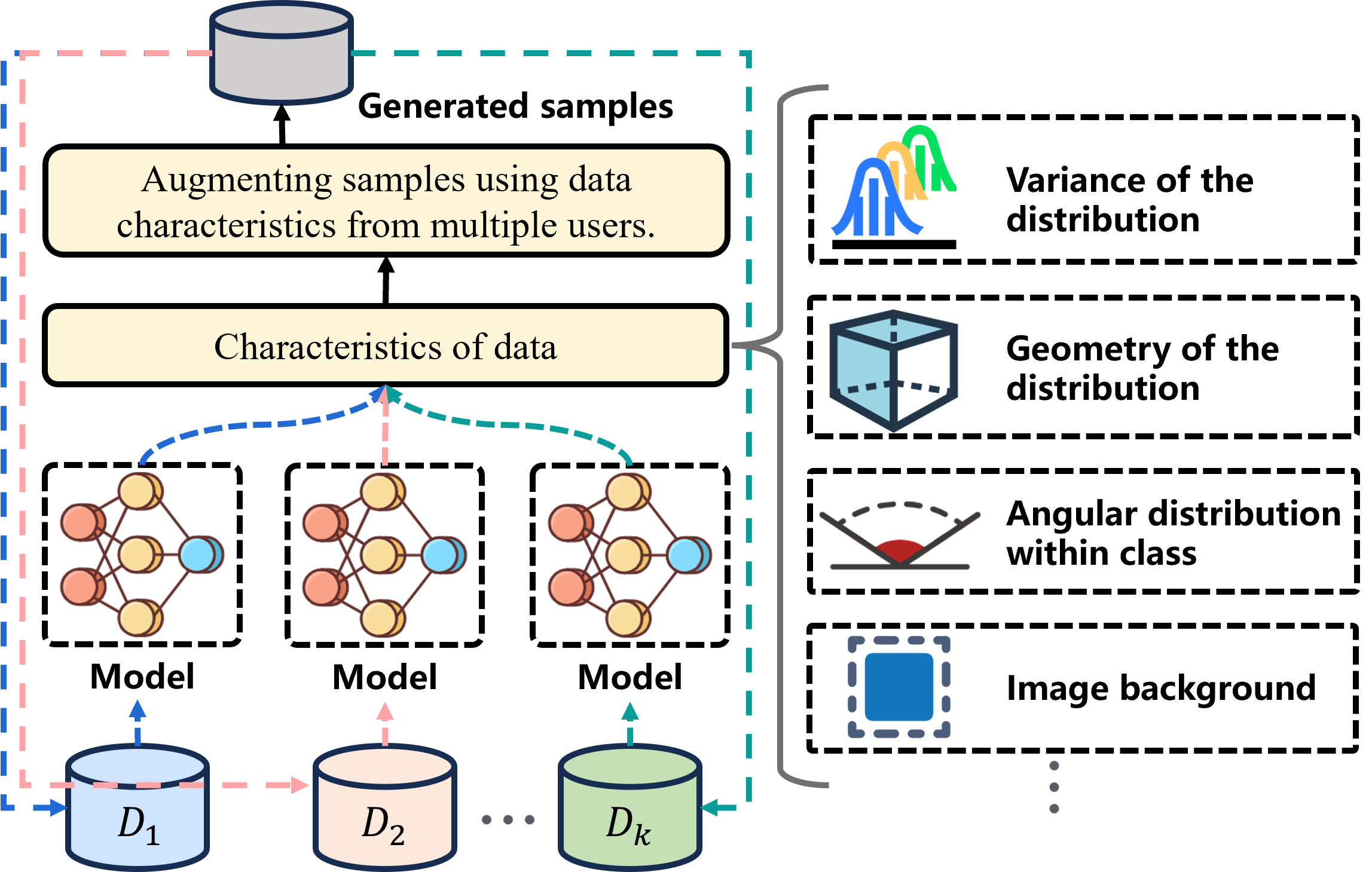}
\vskip -0.05in
\caption{A federated learning framework for data generation. $D_k$ represents data from the $k$-th user. Initially, specialized feature extraction models or certain metrics are used to quantify the characteristics of each user's data. These characteristics are then aggregated, and existing methods for information augmentation or generative models are employed to create new samples. Data characteristics can range from simple color distributions to more complex data distribution variances, geometric shapes, and angle distributions.}
\label{fig9}
\vskip -0.05in
\end{figure}

The proposal of a federated learning framework for dataset augmentation is feasible, allowing users to avoid direct data sharing. As depicted in Fig.\ref{fig9}, a common model is used to quantify the characteristics of each user's data, such as statistical properties of their distribution \cite{paper63,paper52,paper26,paper74}. Subsequently, these quantified characteristics are utilized to augment the data for each user \cite{paper107,paper109}, resulting in an improved common dataset. To implement the aforementioned process, \textbf{two crucial issues must be addressed}: (1) which data characteristics to quantify, i.e., which characteristics contribute to efficient data augmentation, and (2) how to effectively employ these quantified characteristics for data augmentation. While existing partial information augmentation methods can tackle the second issue, deeper research is still required for the first issue.

\subsection{Data Evaluation}
\label{sec5.2}

The purpose of data evaluation is to quantify the quality of a dataset, such as whether the number of samples per category is balanced, whether the semantic scale \cite{paper29} of categories is balanced, and whether there are incorrectly labeled samples in the dataset, among other factors. A thorough dataset evaluation helps address issues effectively and facilitates the initial cleaning of the dataset.

\subsubsection{Data quality}

Poor-quality samples can potentially hurt the performance of a model, such as images with excessive noise or heavily obscured content. While appropriate noise can enhance the robustness of artificial intelligence models, excessive noise can degrade model performance \cite{paper75}. Therefore, \textbf{an important task is to quantify the beneficial level of noise in different tasks and scenarios}. Furthermore, detecting heavily contaminated or heavily obscured images contributes to data cleaning. Some approaches involve removing low-quality samples from long-tailed datasets and selecting balanced subsets, which have shown initial success \cite{paper64,paper77,paper110}. In addition to cleaning low-quality samples from human perception, filtering out useless samples from a model's perspective is also crucial. Recent research \cite{paper11,paper112} has shown that when sampling along the orthogonal directions of data manifolds, models can still recognize these images effectively, even when they are no longer discernible to the human eye. This evidence highlights a significant disparity between the model's recognition capabilities and human visual perception. Shapley scores \cite{paper76} are obtained by observing changes in the model's performance when different samples are removed, providing insight into the importance of each sample from the model's perspective. In summary, both human-perceived data quality assessment and model-perceived data quality assessment should be developed and applied concurrently.

\subsubsection{Detecting mislabelled samples}

The labor-intensive data annotation for complex tasks is prone to errors, such as in common datasets like ImageNet where over $3\%$ of samples are mislabeled \cite{paper78}. Conflicts between samples and labels can disrupt the learning process of artificial intelligence models. Therefore, both bias correction and the removal of mislabeled samples can significantly enhance the performance of long-tailed recognition models \cite{paper79,paper80,paper93}.
Recently, ObjectLab \cite{paper81} introduced a framework for automatically identifying and correcting mislabeled images in object detection, significantly improving the performance of Detectron-X101 and Faster-RCNN on the COCO dataset. \cite{paper82} proposed a qualitative label estimation method for semantic segmentation data using arbitrary models, with lower scores indicating a higher likelihood of mislabeled images. \cite{paper64} discovered that the presence of mislabeled samples in tail classes can be detrimental to model performance. Hence, there is an urgent need for methods to detect mislabeled data in long-tailed scenarios. \textbf{We summarize the following experimental observations, which could facilitate research on mislabel detection in long-tailed scenarios:}

\begin{itemize}
\item[(1)] During the deep neural network learning process, samples with mislabeled are more likely to be forgotten \cite{paper83}.
\item[(2)] Research on unbalanced and noisy CIFAR-10 datasets suggests that noisy samples in tail classes often lead to larger losses, while clean samples in head classes produce the smallest losses \cite{paper84}.
\item[(3)] \cite{paper80} found that the magnitude of loss or classification confidence cannot distinguish between clean and noisy samples in tail classes, which contradicts the findings of (2). Therefore, further investigation is beneficial.
\item[(4)] \cite{paper64} found that even if it is impossible to differentiate between noise and clean samples through instantaneous loss, distinction can be made by observing the loss curves of noisy and clean samples.
\end{itemize}

\subsubsection{Various Measures for Data Imbalance}

\begin{figure}[h]
\centering
\includegraphics[width=3.5in]{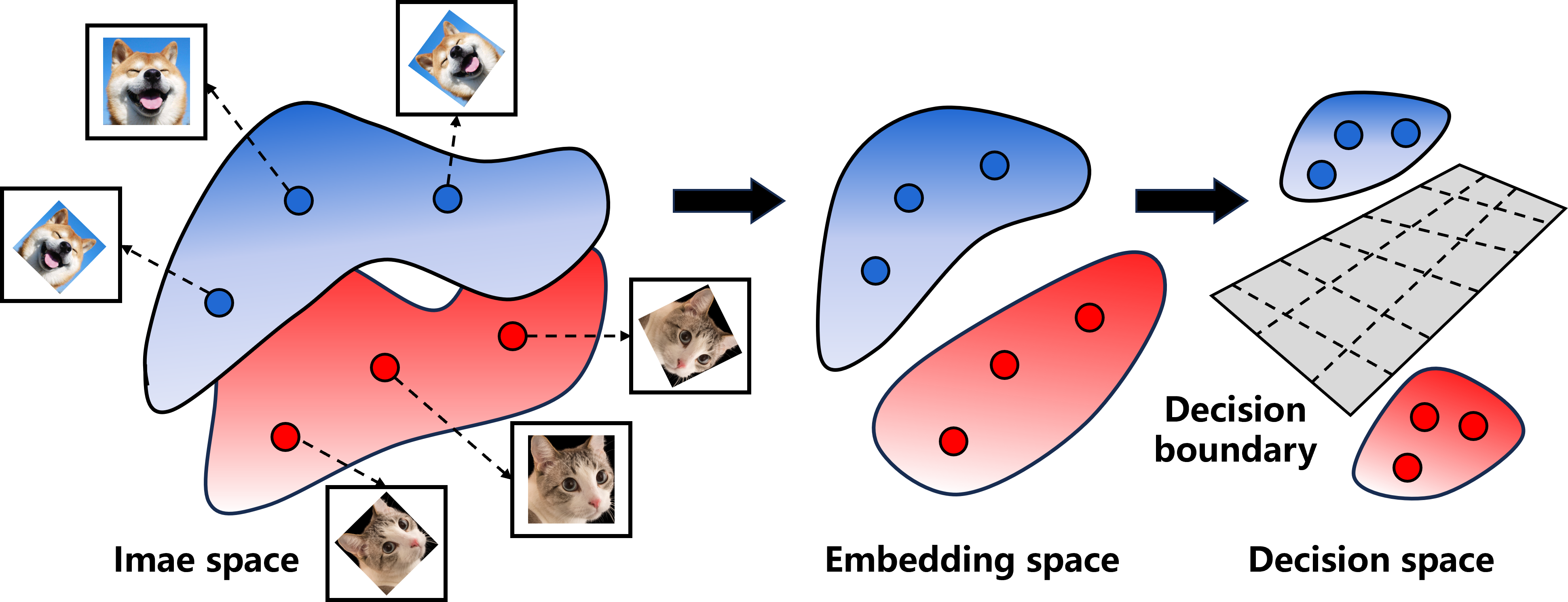}
\vskip -0.05in
\caption{Changes in the geometry of data manifolds as they are transformed in a deep neural network. The classification process of the data includes untangling the manifolds from each other and separating the different manifolds.}
\label{fig7}
\end{figure}

Measuring the degree of imbalance in a dataset helps in targeted improvements during the data augmentation phase. In much of the past research, sample quantities have been used to measure dataset imbalance, with categories having fewer samples referred to as the ``tail'' class. This implies that most research assumes a positive correlation between the model's performance in each category and the number of samples. However, recent studies \cite{paper37,paper36} have shown that models do not necessarily perform poorly on classes with fewer samples, leading to the proposal of non-sample quantity-based imbalance metrics.
For example, semantic scale imbalance \cite{paper29} measures diversity within each category and uses the distribution of diversity to define head and tail classes. The curvature of category-specific data manifolds has also been found to be associated with model bias \cite{paper31}. Therefore, in addition to assessing differences in the number of samples per category, it is necessary to evaluate imbalance under other metrics to systematically improve dataset bias.
For instance, a category may have fewer samples, but its semantic scale (i.e., feature diversity) is not necessarily low compared to other categories. This indicates that the samples in this category provide good coverage of real-world scenarios, and basic data augmentation techniques (such as image flipping or Mixup \cite{paper85}) can be used to augment the samples.
\cite{paper29} and \cite{paper31} open up geometric perspectives for analyzing model bias. As shown in Fig.\ref{fig7}, it treats the classification task as a process of decoupling and separating data manifolds, thereby investigating how the geometric properties of data manifolds affect model bias.
In the future, an attempt can be made to explore the relationship between the intrinsic dimension of the data manifold, the interlayer curvature gap, and the learning difficulty.

During the data evaluation phase, we initially conducted preliminary cleaning by removing some low-quality data from the long-tailed dataset. Additionally, we performed a quantitative assessment of various types of data imbalances within the dataset. These imbalance metrics will be utilized to guide the effective application of information augmentation methods.

\subsection{Selecting Representative Data for Visualization}
\label{sec5.3}

Data visualization is integrated throughout the entire data processing framework. Visualizing the results of each processing stage facilitates subjective assessments by researchers or users. In certain contexts with stringent security requirements, user expertise is essential for evaluating data processing results because relying solely on algorithms or models to optimize the dataset cannot always guarantee safety and reliability.
The primary task of visualization is to select a relatively small subset from a large number of samples, which should effectively represent the range of data distribution. Users can assess the quality of data processing by simply observing this subset, without the need to examine the entire dataset. However, random sampling of subsets cannot guarantee comprehensive coverage of the data distribution in every instance, necessitating the development of new data sampling methods. The determinant point process \cite{paper111} is a potential solution, as it offers a more uniform sampling approach compared to random sampling.

\subsection{Enhancing the Dataset}
\label{sec5.4}

After evaluating the dataset, we performed a cleaning process on some low-quality samples to obtain a clean dataset. Failing to conduct data cleaning before applying information augmentation could potentially amplify errors, thus reducing the dataset's quality.


The dataset may exhibit imbalances across multiple metrics, necessitating targeted improvements and enhancements. Traditional data augmentation methods such as image flipping, cropping, Mixup, Cutout \cite{paper86}, CutMix \cite{paper87}, and others struggle to generate guided out-of-distribution samples. However, these methods do not require additional knowledge guidance and are straightforward to implement. On the other hand, knowledge transfer-based approaches like FDC \cite{paper63}, LEAP \cite{paper26}, OFA \cite{paper5}, and others explore knowledge that can be used to recover the real distribution of tail classes, such as variance \cite{paper74, paper63}, geometric shape of data distribution \cite{paper24}, intra-class angle distribution \cite{paper52}, and image backgrounds \cite{paper33}. When a class has a scarce number of samples and significant diversity is lacking (measured using semantic scales), traditional data augmentation methods struggle to be effective. Thus, introducing additional knowledge to expand the class distribution becomes necessary. In another scenario, even though a class has very few samples, these samples essentially cover practical application scenarios, making traditional data augmentation methods sufficient for achieving good results \cite{paper66}. In practical applications, situations where samples cannot encompass real-world scenarios are more complex, requiring further research into knowledge that can be used to recover the true distribution of tail classes.

\begin{figure}[h]
\centering
\includegraphics[width=3.5in]{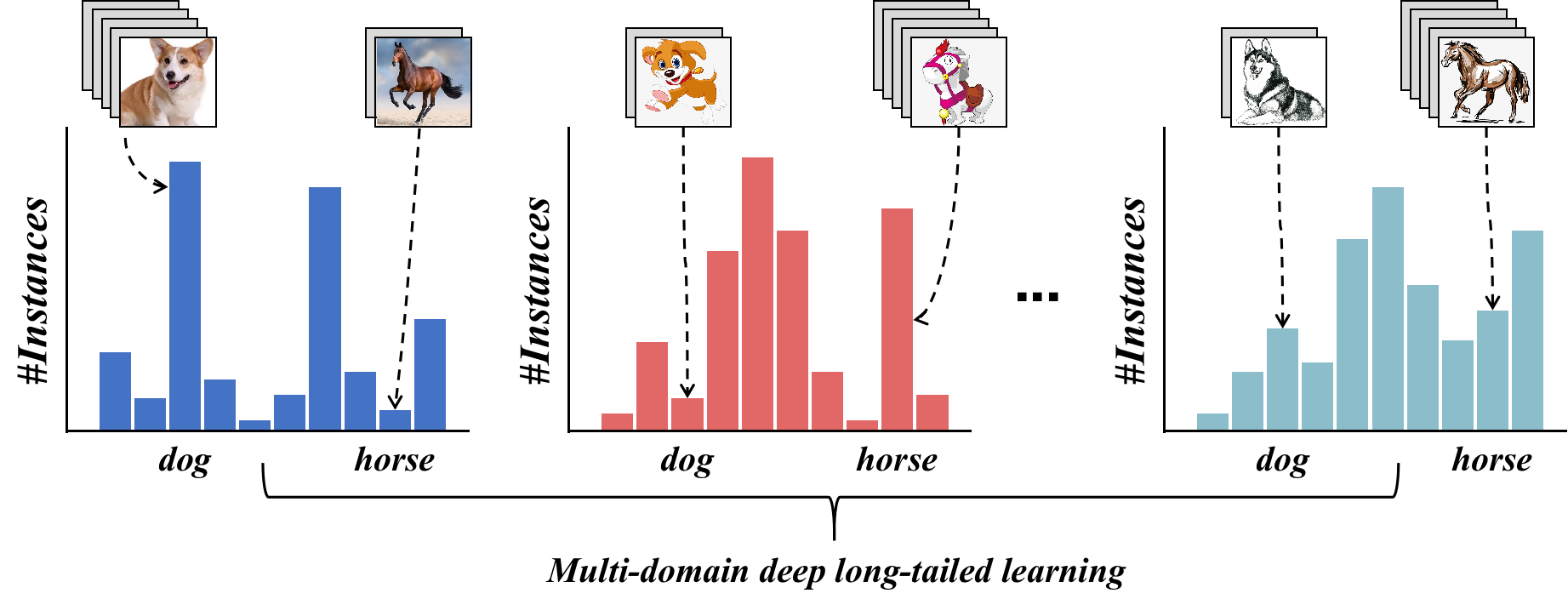}
\vskip -0.05in
\caption{Creating a balanced dataset with an equal number of samples by utilizing images from multiple domains. For instance, using cartoon or sketch images to supplement the tail classes in the real-world dataset.}
\label{fig8}
\end{figure}

The virtual images generated by models can be used to enrich the data of tail classes \cite{paper88,paper89,paper90}, but this approach has two main limitations \cite{paper66}. Firstly, there is always a gap between synthetic images and real ones, and they cannot fully replace real-world scenarios. Secondly, we need to determine what content should be included in synthetic images, but currently, we lack sufficient prior knowledge to guide the generative models \cite{paper90}. With the rapid development of generative models, synthetic images are gradually becoming closer to the real world and will be used more extensively for data augmentation. 

Collecting images from multiple domains for data augmentation has been found to be an effective measure \cite{paper91}. Existing long-tailed datasets all originate from the same domain, but natural data can come from various different domains. Certain domains may have an abundance of samples for tail classes that are scarce in another domain. As illustrated in Fig.\ref{fig8}, using cartoon/sketch images to augment classes lacking samples in the real-world dataset helps alleviate the long-tailed problem. However, this approach requires aligning the image features from multiple domains.

Unlike adding noise to images, recent research \cite{paper53} has found that using pure noise images as augmentation samples can significantly improve the model's performance on tail classes. Reasonably utilizing noise with a positive impact is a potential direction \cite{paper92}, but \textbf{it requires addressing two key problems: the choice of noise type and the determination of noise level.}

\subsection{Learning Strategies and Data Maintenance}
\label{sec5.5}

The data-centric long-tail learning framework not only provides users with high-quality dataset feedback but also offers recommendations for appropriate training strategies. For instance, it can provide users with feedback on the assessment of sample importance, helping them distinguish between challenging and straightforward samples. This aids users in engaging in active learning, self-paced learning, or comparative learning.
The application scenarios of AI models may change over time, such as equipment replacements or environmental renovations. Therefore, it is essential to continuously monitor whether the newly generated sample distribution deviates significantly from the distribution of the training samples. In the case of substantial deviation, alerts should be sent to model developers, indicating the need for model retraining. Additionally, in situations with a substantial volume of continuously generated data, algorithmic selection of data for dataset creation becomes crucial, especially when filtering tail-class samples.

\section{Summary and outlook}
\label{Summary and outlook}

In most real-world applications, data exhibits a long-tail distribution, resulting in biases in trained models. To ensure that models perform well across various scenarios, model-centric approaches like re-weighted loss, transfer learning, ensemble learning, etc., have been widely proposed. However, as foundational models advance rapidly, model-centric approaches have gradually shown limitations.
Returning to the root cause, directly improving the data is the essence of addressing the long-tailed problem. Past research has often focused solely on how to augment the tail classes, overlooking the fact that enhancing the dataset is a systemic endeavor. In this work, we introduce a data-centric long-tailed learning framework comprising a set of foundational tasks. Additionally, we provide a tool, namely Feature Diversity Gain, to enhance the effectiveness of information augmentation techniques. In the future, we aim to propose data-centric long-tail recognition benchmark datasets, driving the practical application of AI systems.

\bibliographystyle{IEEEtran}
\bibliography{TKDE2023}

\end{document}